\documentclass{clv2}

\usepackage{lscape}
\usepackage{epsfig}
\usepackage{float}
\usepackage{longtable}
\usepackage{url}
\usepackage{xtab}
\usepackage{tabularx}
\usepackage{makeidx}
\usepackage{graphicx}
\usepackage{epsf}
 \usepackage{mathpazo}
\usepackage{array}
\usepackage{lscape}

\makeatletter
\newcommand\figcaption{\def\@captype{figure}\caption}
\makeatother

\issue{1}{1}{2012}

\dochead{}

\runningtitle{Computing Lexical Contrast}

\runningauthor{Mohammad, Dorr, Hirst, Turney}
\historydates{Submission received: 14 January 2010.\\
Revised submission received: 26 June 2012.\\
Accepted for publication: 16 July 2012.}

\begin{document}

\title{Computing Lexical Contrast}

 \author{Saif M. Mohammad\thanks{National Research Council Canada. E-mail: saif.mohammad@nrc-cnrc.gc.ca}}
 \affil{National Research Council Canada}

 \author{Bonnie J. Dorr\thanks{Department of Computer Science and Institute of Advanced Computer Studies, University of Maryland. E-mail: bonnie@umiacs.umd.edu}}
 \affil{University of Maryland}

\author{Graeme Hirst\thanks{Department of Computer Science, University of Toronto. E-mail: gh@cs.toronto.edu}}
\affil{University of Toronto}

 \author{Peter D. Turney\thanks{National Research Council Canada. E-mail: peter.turney@nrc-cnrc.gc.ca}}
 \affil{National Research Council Canada}

\maketitle

\begin{abstract}
Knowing the degree of semantic contrast between words
has widespread application in natural language processing, including machine translation,
information retrieval, and dialogue systems.
 Manually-created lexicons focus on opposites, such as {\rm hot} and {\rm cold}.
Opposites are of many kinds such as antipodals, complementaries, and gradable.
However, existing lexicons often do not classify opposites into the different kinds.
 They also do not explicitly list word pairs that are not opposites but yet have some degree of contrast in meaning, such as
{\rm warm} and {\rm cold} or {\rm tropical} and  {\rm freezing}. 
We propose an automatic method to identify contrasting word pairs that is based on the 
hypothesis that 
 if a pair of words, $A$ and $B$, are contrasting, then there is a pair of opposites, $C$ and $D$, such that $A$ and $C$ are strongly related and $B$ and $D$ are strongly related.
(For example, there exists the pair of opposites {\rm hot} and {\rm cold} such that {\rm tropical} is related to {\rm hot,} and {\rm freezing} is related to {\rm cold}.) 
 We will call this the contrast hypothesis. 

We begin with a large crowdsourcing experiment to determine
the amount of human agreement on the concept of oppositeness and its 
different kinds. In the process, we flesh out key features of different
kinds of opposites.
We then present an automatic and empirical measure of lexical contrast that
relies on the contrast hypothesis, corpus statistics, and the structure of a {\it Roget}-like thesaurus.
We show how using four different datasets, we evaluated our approach on two different tasks, solving
 ``most contrasting word'' questions and distinguishing synonyms from opposites.
The results are analyzed across four parts of speech and across five different kinds of opposites.
We show that the proposed measure of lexical contrast  obtains
high precision and large coverage, outperforming existing methods.
\end{abstract}

\section{Introduction}
Native speakers of a language intuitively recognize different {\it degrees
of lexical contrast}---for example most people will agree that {\it hot} and {\it cold}
have a higher degree of contrast than {\it cold} and {\it lukewarm},
and {\it cold} and {\it lukewarm} have a higher degree of contrast than
{\it penguin} and {\it clown}.
Automatically determining the degree of contrast between words has
many uses, including:
\begin{itemize}
\item Detecting and generating paraphrases \cite{MartonWMT11}
({\it The dementors {\bf \em caught} Sirius Black} /
{\it Black could {\bf \em not  escape} the dementors}).
\item Detecting certain types of contradictions \cite{MarneffeRM08,Voorhees08}
({\it Kyoto has a predominantly {\bf \em wet} climate} /
{\it It is mostly {\bf \em dry} in Kyoto}).
This is in turn useful in
effectively re-ranking target language hypotheses in machine translation,
and for re-ranking query responses in information retrieval.
\item Understanding discourse structure and improving dialogue systems.
Opposites often indicate the discourse relation of contrast \cite{MarcuE02}.
\item Detecting humor \cite{MihalceaS05}.
Satire and jokes tend to have contradictions and oxymorons.
\item Distinguishing near-synonyms from word pairs that are semantically contrasting
in automatically created distributional thesauri.
Measures of distributional similarity typically fail to do so.
\end{itemize}
\noindent  Detecting lexical contrast is not sufficient by itself
 to solve most of these problems, but it is a crucial
 component.

Lexicons of pairs of words that native speakers consider opposites have been
created for certain languages, but their coverage is limited.
Opposites are of many kinds such as antipodals, complementaries, and gradable (summarized ahead in Section 3).
However, existing lexicons often do not classify opposites into the different kinds.
Further, the terminology is inconsistent across different sources. For example, 
Cruse \shortcite{Cruse86} defines {\it antonyms} as gradable adjectives that are opposite in meaning,
whereas the WordNet antonymy link connects some verb pairs, noun pairs, and adverb pairs too.
In this paper, we will follow Cruse's terminology, and we will refer to word pairs
connected by WordNet's antonymy link as opposites, unless referring specifically to gradable adjectival pairs.

Manually created lexicons also do not explicitly list word pairs that are not opposites but yet have some degree of contrast in meaning, such as
{\it warm} and {\it cold} or {\it tropical} and {\it cold}.
Further, 
contrasting word pairs far outnumber those that are commonly considered opposites.
In our own experiments described later in this paper, we find that more than 90\% of the contrasting pairs
in GRE ``most contrasting word'' questions are not listed as antonyms in WordNet.
 We should not infer from this that WordNet or any other lexicographic resource is a poor source for detecting opposites, 
but rather that identifying the large number of contrasting word pairs requires further computation,
possibly relying on other semantic relations stored in the lexicographic resource.

Even though a number of computational approaches have
been proposed for semantic closeness \cite{BudanitskyH06,Curran04}, and some for
hypernymy--hyponymy \cite{Hearst92}, measures of lexical contrast have been less successful.
To some extent, this is because lexical contrast
is not as well understood as other classical lexical-semantic
relations.  

 Over the years, many definitions of semantic contrast and opposites have been proposed
 by linguists \cite{Cruse86,LehrerL82}, cognitive scientists
 \cite{Kagan84}, psycholinguists \cite{Deese65}, and lexicographers
 \cite{Egan84},
 which differ from each other in various respects.
Cruse \shortcite{Cruse86} observes that even though people have a robust
intuition of opposites, ``the overall class is not a well-defined one.''
He points out that a defining feature of opposites is that they tend
to have many common properties, but
differ saliently along one dimension of meaning.
We will refer to this semantic dimension as the {\it dimension of opposition}.  
For example, {\it giant} and {\it dwarf} are both living beings, they both eat,
they both walk, they are are both capable of thinking, and so on.
However, they are most saliently different along the dimension of height.
Cruse also points out that sometimes it is difficult to identify or articulate the dimension
of opposition (for example, {\it city--farm}).

Another way to define opposites is that they are word pairs with a ``binary
incompatible relation'' \cite{Kempson77}.
That is to say that one member entails the absence of the other, and given one member,
the identity of the other member is obvious. Thus, {\it night} and {\it day}
are good examples of opposites because {\it night} is best paraphrased by {\it not day},
rather than the negation of any other term. On the other hand, {\it blue} and {\it yellow}
make poor opposites because even though they are incompatible, they do not have an
obvious binary relation such that {\it blue} is understood to be a negation of {\it yellow}.
It should be noted that there is a relation between binary incompatibility and
difference along just one dimension of meaning.

For this paper, we define {\it opposites} to be term pairs that clearly satisfy either the property of binary incompatibility or the property of salient difference across a dimension of meaning.
However, word pairs may satisfy the two properties to different degrees. 
We will refer to all word pairs that satisfy either of the two properties to some degree as {\it contrasting}.
For example, {\it daylight} and {\it darkness} are very different along the dimension
of light, and they satisfy the binary incompatibity property to some degree, but not as strongly
as {\it day} and {\it night}.
Thus we will consider both {\it daylight} and {\it darkness} as well as {\it day} and {\it night} as semantically contrasting pairs (the former pair less so than the latter),
but only {\it day} and {\it night} as opposites.
Even though there are subtle differences in the meanings of the terms {\it contrasting, opposite,} and {\it antonym}, they have often been used interchangeably in the literature,
dictionaries, and common parlance. Thus, we list below what we use these terms to mean in this paper:
\vspace*{-2mm}
\begin{itemize}
\item {\it Opposites} are word pairs that have a strong binary incompatibility relation with each other and/or are saliently different across a dimension of meaning.
\vspace*{-1mm}
\item {\it Contrasting word pairs} are word pairs that have some non-zero degree of binary incompatibility and/or have some non-zero difference across a dimension of meaning.
Thus, all opposites are contrasting, but not all contrasting pairs are opposites.
\vspace*{-1mm}
\item {\it Antonyms} are opposites that are also gradable adjectives.\footnote{We follow Cruse's \shortcite{Cruse86} definition for antonyms. However, the WordNet antonymy link also connects some verb pairs, noun pairs, and adverb pairs.}
\vspace*{-2mm}
\end{itemize}

In this paper, we present 
an automatic method to identify contrasting word pairs that is based on the following hypothesis: 
\vspace{-1mm}
\begin{quote}
{\it Contrast Hypothesis}: If a pair of words, $A$ and $B$, are contrasting, then there is a pair of opposites, $C$ and $D$, such that $A$ and $C$ are strongly related and $B$ and $D$ are strongly related.
\end{quote}
\vspace{-2mm}
For example, there exists the pair of opposites {\it night} and {\it day} such that {\it darkness} is related to {\it night}, and {\it daylight} is related to {\it day}.
 We then determine the degree of contrast between two words using the hypothesis stated below:
\begin{quote}
{\it Degree of Contrast Hypothesis}: If a pair of words, $A$ and $B$, are contrasting, then their
degree of contrast is proportional to their tendency to co-occur in a large corpus.\\
\vspace{-6mm}
\end{quote}
For example, consider the contrasting word pairs {\it top--low} and {\it top--down}, since
{\it top} and {\it down} occur together much more often than {\it top} and {\it low},
our method concludes that the pair {\it top--down} has a higher degree of lexical contrast than the pair {\it top--low}.
The degree of contrast hypothesis is inspired by the idea that opposites tend to co-occur more often than chance \cite{CharlesM89,Fellbaum95}.
Murphy and Andrew~\shortcite{MurphyA93} claim that
this is because together opposites convey contrast well, which is rhetorically useful.
Thus we hypothesize that the higher the degree of contrast between two words, the higher the tendency of people to use them together.

Since opposites are a key component of our method, we begin by first understanding different kinds of opposites (Sections~\ref{sec:kindsofopps}).
Then we describe a crowdsourced project on the annotation of opposites into different kinds (Section \ref{sec:crowd}).
 In Section~\ref{sec:cohyp}, we examine whether opposites and other highly contrasting word pairs occur together in text more often than randomly chosen word pairs.
This experiment is crucial to the degree of contrast hypothesis since 
if it is true, then we should find that highly contrasting pairs are used together much more often than randomly chosen word pairs.
Section~\ref{sec:disthyp} examines this question.
Section~\ref{sec:method} presents our method to automatically
compute the degree of contrast between word pairs by
relying on the contrast hypothesis, the degree of contrast hypothesis, seed opposites, and the structure of a {\it Roget}-like thesaurus.
(This method was first described in Mohammad et al.\@ \shortcite{MohammadDH08}.)
Finally we present
experiments that evaluate various aspects of the automatic method (Section 7).
Following is a summary of the key research questions
addressed by this paper:\\


{\bf 1. On the kinds of opposites:} \\

\vspace*{-2mm}
{\addtolength{\leftskip}{13mm} \addtolength{\rightskip}{7mm}
\noindent {\bf Research questions:}
How good are humans at identifying different kinds of opposites? 
Can certain terms pairs belong to more than one kind of opposite?\\

\vspace*{-2mm}
\noindent {\bf Experiment:}
In Sections 3 and 4, we describe how we designed a questionnaire to acquire annotations about opposites. 
Since the annotations are done by crowdsourcing, and there is no control over
the educational background of the annotators, we devote
extra effort to make sure that the questions are phrased in a simple, yet clear manner.
We deploy a quality control method that uses a
word-choice question to automatically identify and discard dubious and outlier annotations. \\

\vspace*{-2mm}
\noindent {\bf Findings:}
We find that humans agree markedly in identifying opposites;
however, there is significant variation in the agreement for different kinds of opposites.
We find that a large number of opposing word pairs have properties pertaining to more than one kind of opposite. \\

}

{\bf 2. On the manifestation of opposites and other highly contrasting pairs in text:}\\

\vspace*{-2mm}
{\addtolength{\leftskip}{13mm} \addtolength{\rightskip}{7mm}
\noindent {\bf Research questions:}
How often do highly contrasting word pairs co-occur in text? How strong is this tendency compared to random word pairs,
and compared to near-synonym word pairs?\\

\vspace*{-2mm}
\noindent {\bf Experiment:} 
Section 5 describes how we compiled sets of highly contrasting word pairs (including opposites), near-synonym pairs, and random word pairs, and determine
the tendency for pairs in each set to co-occur in a corpus.\\

\vspace*{-2mm}
\noindent {\bf Findings:}
\noindent Highly contrasting word pairs co-occur significantly more often than both the random word pairs set and also the near-synonyms set.
We also find that the average distributional similarity of highly contrasting word pairs is higher than that of synonymous words.
However, the standard deviations of the distributions for the high-contrast set and the synonyms set are large and so the tendency to co-occur 
is not sufficient to distinguish highly contrasting word pairs from near-synonymous pairs.\\

}

{\bf 3. On an automatic method for computing lexical contrast:}\\

\vspace*{-2mm}
{\addtolength{\leftskip}{13mm} \addtolength{\rightskip}{7mm}
\noindent {\bf Research questions:}
 How can the contrast hypothesis and the degree of contrast hypothesis be used to develop an automatic method for identifying contrasting word pairs?
How can we automatically generate the list of opposites, which are needed as input for a method relying on the contrast hypothesis?\\

\vspace*{-2mm}
\noindent {\bf Proposed Method:}
Section~\ref{sec:method} describes an empirical method for determining the degree of contrast between two words by using the contrast hypothesis, 
 the degree of contrast hypothesis, the structure of a thesaurus, and seed opposite pairs.
The use of affixes to generate seed opposite pairs is also described.
(This method was first proposed in Mohammad et al.\@ \shortcite{MohammadDH08}.)\\

}

{\bf 4. On the evaluation of automatic methods of contrast:}\\

\vspace*{-2mm}
{\addtolength{\leftskip}{13mm} \addtolength{\rightskip}{7mm}
\noindent {\bf Research questions:}
How accurate are automatic methods at identifying whether one word pair has a higher degree of contrast than another? 
 What is the accuracy of this method in detecting opposites (a notable subset of the contrasting pairs)?
How does this accuracy vary for different kinds of opposites?\footnote{Note that though linguists have classified opposites into different kinds,
we know of no work doing so for contrasts more generally. Thus this particular analysis must be restricted to opposites alone.}
How easy is it for automatic methods to distinguish between opposites and synonyms?
How does the proposed method perform when compared to other automatic methods?\\

\vspace*{-2mm}
\noindent {\bf Experiments:}
We conduct three experiments (described in Sections 7.1, 7.2, and 7.3) involving three different datasets and two tasks
to answer these questions. We compare performance of our method with methods proposed by Lin et al.\@ \shortcite{LinZQZ03} and Turney \shortcite{Turney08}.
We automatically generate a new set of 1296 ``most contrasting word'' questions to evaluate performance of our method on five different kinds of opposites and across four parts of speech. 
(The evaluation described in Section 7.1 was first described in Mohammad et al.\@ \shortcite{MohammadDH08}.)\\

\vspace*{-2mm}
\noindent {\bf Findings:}
We find that the proposed measure of lexical contrast  obtains
high precision and large coverage, outperforming existing methods.
Our method performs best on gradable pairs, antipodal pairs, and complementary pairs, but poorly
on disjoint opposite pairs.
Among different parts of speech, the method performs best on noun pairs, and relatively worse on verb pairs.\\

}

\noindent  All of the data created and compiled as part of this research is
summarized in Table~\ref{tab:data} (Section~\ref{sec:conclusion}), and is available for download.\footnote{\url{http://www.purl.org/net/saif.mohammad/research}}

\section{Related work}
\label{sec:relwork}

Charles and Miller \shortcite{CharlesM89} proposed that opposites
occur together in a sentence more often than chance.  This is known as
the {\it co-occurrence hypothesis}.  
Paradis et al.\@ \shortcite{Paradis09} describe further experiments to show how canonical opposites tend to have high textual co-occurrence.
Justeson and Katz \shortcite{JustesonK91} gave evidence in support of the hypothesis using 35
prototypical opposites (from an original set of 39 opposites compiled
by Deese \shortcite{Deese65}) and also with an additional 22 frequent opposites.
They also showed that opposites tend to occur in parallel syntactic constructions.
All of these pairs were adjectives.  Fellbaum \shortcite{Fellbaum95} conducted
similar experiments on 47 noun, verb, adjective, and adverb pairs (noun--noun, noun--verb, noun--adjective, verb--adverb and so on)
pertaining to 18 concepts (for example, {\it lose(v)--gain(n)} and {\it loss(n)--gain(n)}, where {\it lose(v)} and {\it loss(n)}
pertain to the concept of ``failing to have/maintain'').
However, non-opposite semantically related words 
also tend to occur together more often than chance.
Thus, separating opposites from these other classes has proven to be difficult.

Some automatic methods of lexical contrast rely on lexical patterns in text.
For example, Lin et al.\@ \shortcite{LinZQZ03} used patterns such as ``from {\it X} to {\it Y}$\,$''
and ``either {\it X} or {\it Y}$\,$''
to separate opposites from distributionally similar pairs.
They evaluated their method on 80 pairs of opposites and 80 pairs of synonyms
taken from the {\it Webster's Collegiate Thesaurus} \cite{Kay88}.       
The evaluation set of 160 word pairs was chosen such that it
included only high-frequency terms. This was necessary to increase
the probability of finding sentences in a corpus where the target pair occurred in one of the chosen patterns.
Lobanova et al.\@ \shortcite{Lobanova10} used a set of Dutch adjective seed pairs
to learn lexical patterns commonly containing opposites. 
The patterns were in turn used to create a larger list of Dutch opposites. The method was
evaluated by comparing entries to Dutch lexical resources and by asking
human judges to determine whether an automatically found pair is indeed an opposite.
Turney \shortcite{Turney08} proposed a supervised method 
for identifying synonyms, opposites, hypernyms, and other
lexical-semantic relations between word pairs. The approach learns patterns
corresponding to different relations.


Harabagiu et al.\@ \shortcite{HarabagiuHL06} detected contrasting word pairs for the purpose of
identifying contradictions by using WordNet chains---synsets connected by
the hypernymy--hyponymy links and exactly one antonymy link.
Lucerto et al.\@ \shortcite{LucertoPJ04} proposed detecting contrasting word pairs using
the number of tokens between two words in text and also cue words such as {\it but},
{\it from}, and {\it and}. Unfortunately, they evaluated their method on only
18 word pairs. 
Neither Harabagiu et al.\@ nor Lucerto et al.\@ determined the degree of contrast between words
and their methods have not been shown to have substantial coverage.

 Schwab et al.\@ \shortcite{SchwabLP02} created an oppositeness vector
 for a target word. The closer this vector is to the context vector of the other target word,
 the more opposite the two target words are. The oppositeness vectors were created
 by first manually identifying possible opposites and then generating suitable vectors for each
 using dictionary definitions.
 The approach was evaluated on only a handful of word pairs.

There is a large amount of work on sentiment analysis and opinion mining aimed at determining
the polarity of words \cite{PangL08}. For example, Pang, Lee, and Vaithyanathan \shortcite{PangLV02} detected
that adjectives such as {\it dazzling, brilliant,} and {\it gripping} cast their qualifying
nouns positively whereas adjectives such as {\it bad, cliched,} and {\it boring}
portray the noun negatively. Many of these gradable adjectives have opposites,
but these approaches, with the exception of that of Hatzivassiloglou and McKeown \shortcite{HatzivassiloglouM97}, did not attempt to determine pairs
of positive and negative polarity words that are opposites. 
Hatzivassiloglou and McKeown \shortcite{HatzivassiloglouM97} proposed a supervised algorithm 
that uses word usage patterns to generate a graph with adjectives as nodes. 
An edge between two nodes indicates either that the two adjectives have the same or opposite polarity. 
A clustering algorithm then partitions the graph into two subgraphs such that the nodes in a subgraph have the same polarity. 
They used this method to create a lexicon of positive and negative words, and argued that
the method could also be used to detect opposites.



\section{The Heterogeneous Nature of Opposites}
\label{sec:kindsofopps}

Opposites, unlike synonyms, can be of different kinds.
Many different classifications have been proposed, one of 
which is given by Cruse
\shortcite{Cruse86} (Chapters 9, 10, and 11).
It consists of complementaries ({\it open--shut, dead--alive}),
antonyms  ({\it long--short, slow--fast}) (further classified
into polar, overlapping, and equipollent opposites),
directional opposites ({\it up--down, north--south}) (further classified into
antipodals, counterparts, and reversives),
relational opposites ({\it husband--wife, predator--prey}),
indirect converses ({\it give--receive, buy--pay}),
congruence variants ({\it huge--little, doctor--patient}), and
pseudo opposites ({\it black--white}).

Various lexical
relations have also received attention at the Educational Testing
Services (ETS), as analogies and ``most contrasting word'' questions are part of
the tests they conduct.  They classify opposites into contradictories
({\it alive--dead, masculine--feminine}), contraries ({\it old--young,
happy-sad}), reverses ({\it attack--defend, buy--sell}), directionals
({\it front--back, left--right}), incompatibles ({\it happy--morbid,
frank--hypocritical}), asymmetric contraries ({\it hot--cool,
dry--moist}), pseudoopposites ({\it popular--shy, right--bad}), and
defectives ({\it default--payment, limp--walk}) \cite{BejarCE91}.

Keeping in mind the meanings and subtle distinctions between
each of these kinds of opposites is not easy even if we provide extensive training to annotators.
Since we crowdsource the annotations, and we know that Turkers prefer to spend their time 
doing the task (and making money) rather than reading lengthy descriptions,
we focused only on five kinds of opposites that we
believed would be easiest to annotate, and which still captured a majority
of the opposites:
\begin{itemize}
\item {\bf Antipodals} ({\it top--bottom, start--finish}): Antipodals are opposites in which
``one term represents an extreme in one direction along some salient axis, while the other term
denotes the corresponding extreme in the other direction'' \cite{Cruse86}.
\item {\bf Complementaries} ({\it open--shut, dead--alive}): The essential characteristic of  a pair of complementaries is that
``between them they exhaustively divide the conceptual domain into two mutually exclusive compartments, so that what does not fall into one
of the compartments must necessarily fall into the other'' \cite{Cruse86}.
\item {\bf Disjoint} ({\it hot--cold}, {\it like--dislike}): Disjoint opposites
are word pairs that occupy non-overlapping regions in the semantic dimension such that
there are regions not covered by either term.  
This set of opposites includes equipollent adjective pairs (for example, {\it hot--cold}) and stative verb pairs (for example, {\it like--dislike}). We refer the reader
to Sections 9.4 and 9.7 of Cruse \shortcite{Cruse86} for details about these sub-kinds of opposites.
\item {\bf Gradable opposites} ({\it long--short, slow--fast}): are adjective-pair or adverb-pair opposites that are gradable,
that is, ``members of the pair denote degrees of some variable property such as length, speed, weight, accuracy, etc'' \cite{Cruse86}.
\item {\bf Reversibles} ({\it rise--fall, enter--exit}): Reversibles are opposite verb pairs such that ``if one member
denotes a change from A to B, its reversive partner denotes a change from B to A'' \cite{Cruse86}.
\end{itemize}
\noindent It should be noted that there is no agreed-upon number of kinds of opposites. Different researchers have proposed various classifications which overlap to a greater or lesser degree. 
It is possible that for a certain application or study one may be interested in a kind of opposite not listed above.

%
%

\section{Crowdsourcing}
\label{sec:crowd}
We used the {\it Amazon Mechanical Turk (AMT)} service to obtain annotations for different kinds of opposites.
We broke the task into small independently solvable units called {\it HITs
(Human Intelligence Tasks)} and uploaded them on the AMT
website.\footnote{\url{https://www.mturk.com/mturk/welcome}}
Each HIT had a set of questions, all of which were to be answered by the same person (a {\it Turker}, in AMT parlance).  
We created HITs for word pairs, taken from WordNet, that we expected to have some degree of contrast in meaning.

 In WordNet, words that are close in meaning are grouped together in a set called a {\it synset}. If one of the words in a synset
is an opposite of another word in a different synset, then the two synsets are called {\it head synsets} and WordNet
records the two words as {\it direct antonyms} 
\cite{GrossFM88}---WordNet regards the terms {\it opposite} and {\it antonym} as synonyms.
Other word pairs across the two head synsets are called  {\it indirect antonyms}.
Since we follow Cruse's definition of antonyms which requires antonyms to be gradable adjectives,
and since WordNet's direct antonyms include noun, verb, and adverb pairs too,
for the rest of the paper, we will refer to WordNet's direct antonyms as {\it direct opposites}
WordNet indirect antonyms as {\it indirect opposites}.
We will refer to the union of both the direct and indirect opposites simply as {\it WordNet opposites}.
Note that the WordNet opposites are highly contrasting term pairs.

We chose as target pairs all the direct or indirect opposites from WordNet
that were also listed in the {\it Macquarie Thesaurus}.
This condition was a mechanism to ignore less-frequent and obscure words, and 
apply our resources on words that are more common. 
Additionally, as we will describe ahead, we use the presence of the words in the thesaurus to help generate Question 1,
which we use for quality control of the annotations.
Table \ref{tab:tgt pairs} gives a breakdown of the 1,556 pairs chosen by part of speech.

\begin{table}[]
 \setlength{\baselineskip}{2\baselineskip} 
\caption{Target word pairs chosen for annotation. Each term was annotated about 8 times.}
\centering
\begin{tabular}{l r}

\hline
	  part of speech	&\# of word pairs\\
\hline
	adverbs		&185\\
	adjectives	&646\\
	nouns		&416\\
	verbs		&309\\
	{\bf all}	&{\bf 1556}\\
\hline
\end{tabular}
\label{tab:tgt pairs}
\end{table}

\begin{table}[t!]
\centering
 \scalebox{0.83}{
\begin{tabular}{l}

\vspace{2mm} 
{\bf Word-pair: musical $\times$ dissonant} \\

{\bf Q1.} Which set of words is most related to the word pair musical:dissonant?\\

 \hspace{6mm} \textbullet \hspace{2mm} useless, surgery, ineffectual, institution \\
 \hspace{6mm} \textbullet \hspace{2mm} sequence, episode, opus, composition \\
 \hspace{6mm} \textbullet \hspace{2mm} youngest, young, youthful, immature \\
\vspace{2mm} 
 \hspace{6mm} \textbullet \hspace{2mm} consequential, important, importance, heavy\\
 
{\bf Q2.} Do musical and dissonant have some contrast in meaning? \\
 \hspace{6mm} \textbullet \hspace{2mm} yes \hspace{6mm} \textbullet \hspace{2mm} no\\

For example, up--down, lukewarm--cold, teacher--student, attack--defend, all have at least\\ 
some degree of contrast in meaning. On the other hand, clown--down, chilly--cold,\\
\vspace{2mm} 
teacher--doctor, and attack--rush DO NOT have contrasting meanings.\\

{\bf Q3.} Some contrasting words are paired together so often that given one we naturally\\
think of the other.  If one of the words in such a pair were replaced with another word of\\
almost the same meaning, it would sound odd. Are musical:dissonant such a pair? \\
 \hspace{6mm} \textbullet \hspace{2mm} yes \hspace{6mm} \textbullet \hspace{2mm} no\\

Examples for ``yes'': tall--short, attack--defend, honest--dishonest, happy--sad.\\ 
\vspace{2mm} 
Examples for ``no'': tall--stocky, attack--protect, honest--liar, happy--morbid.\\

{\bf Q5.} Do musical and dissonant represent two ends or extremes? \\
\hspace{6mm} \textbullet \hspace{2mm} yes \hspace{6mm} \textbullet \hspace{2mm} no\\

Examples for ``yes'': top--bottom, basement--attic, always--never, all--none, start--finish.\\ 
Examples for ``no'': hot--cold (boiling refers to more warmth than hot and freezing refers\\
to less warmth than cold), teacher--student (there is no such thing as more or less teacher\\
\vspace{2mm} 
and more or less student), always--sometimes (never is fewer times than sometimes).\\

{\bf Q6.} If something is musical, would you assume it is not dissonant, and vice versa?\\
In other words, would it be unusual for something to be both musical and dissonant?\\
\hspace{6mm} \textbullet \hspace{2mm} yes \hspace{6mm} \textbullet \hspace{2mm} no\\

Examples for ``yes'': happy--sad, happy--morbid, vigilant--careless, slow--stationary.\\ 
\vspace{2mm} 
Examples for ``no'': happy--calm, stationary--still, vigilant--careful, honest--truthful.\\

{\bf Q7.} If something or someone could possibly be either musical or dissonant, is it\\
necessary that it must be either musical or dissonant? In other words, is it true that for\\
things that can be musical or dissonant, there is no third possible state, except perhaps\\
under highly unusual circumstances? \\
\hspace{6mm} \textbullet \hspace{2mm} yes \hspace{6mm} \textbullet \hspace{2mm} no\\

Examples for ``yes'': partial--impartial, true--false, mortal--immortal.\\ 
Examples for ``no'': hot--cold (an object can be at room temperature is neither hot nor cold),\\
\vspace{2mm} 
tall--short (a person can be of medium or average height).\\

{\bf Q8.} In a typical situation, if two things or two people are musical, then can one be more\\
musical than the other? \\
\hspace{6mm} \textbullet \hspace{2mm} yes \hspace{6mm} \textbullet \hspace{2mm} no\\

Examples for ``yes'': quick, exhausting, loving, costly.\\ 
\vspace{2mm} 
Examples for ``no'': dead, pregnant, unique, existent.\\ 

{\bf Q9.} In a typical situation, if two things or two people are dissonant, can one be more\\
dissonant than the other? \\
\hspace{6mm} \textbullet \hspace{2mm} yes \hspace{6mm} \textbullet \hspace{2mm} no\\

Examples for ``yes'': quick, exhausting, loving, costly, beautiful.\\ 
\vspace{2mm} 
Examples for ``no'': dead, pregnant, unique, existent, perfect, absolute.\\ 

\end{tabular}
 }
\figcaption{Example HIT: Adjective pairs questionnaire.}
\label{fig:HIT}
\flushleft
{\footnotesize Note: Perhaps ``musical $\times$ dissonant'' might be better written as ``musical versus dissonant'', but we have\\
\vspace*{-1mm}
kept ``$\times$'' here to show the reader exactly what the Turkers were given.}

{\footnotesize Note: Q4 is not shown here, but can be seen in the online version of the questionnaire. It was an exploratory\\
\vspace*{-1mm}
question, and it was not multiple choice. Q4's responses have not been analyzed.}
\end{table}

\begin{table}[]
\centering
 {\small
\begin{tabular}{l}

{\bf Word-pair: enabling $\times$ disabling} \\
\\
{\bf Q10.} In a typical situation, do the sequence of actions disabling and then enabling bring someone\\
or something back to the original state, AND do the sequence of actions enabling and disabling\\
also bring someone or something back to the original state? \\
\hspace{6mm} \textbullet \hspace{2mm} yes, both ways: the transition back to the initial state makes much sense in both sequences.\\
\hspace{6mm} \textbullet \hspace{2mm} yes, but only one way: the transition back to the original state makes much more sense one\\
\hspace{12mm} way, than the other way.\\
\hspace{6mm} \textbullet \hspace{2mm} none of the above\\

Examples for ``yes, both ways'': enter--exit, dress--undress, tie--untie, appear--disappear.\\ 
Examples for ``yes, but only one way'': live--die, create--destroy, damage--repair, kill--resurrect.\\ 
Examples for ``none of the above'': leave--exit, teach--learn,  attack--defend (attacking and then\\
defending does not bring one back to the original state).\\

\end{tabular}
 }
\figcaption{Additional question in the questionnaire for verbs.}
\label{fig:vquest}
\end{table}

Since we do not have any control over the educational background of the annotators, we made
efforts to phrase questions about the kinds of opposites in a simple and clear manner.
Therefore we avoided definitions and long instructions in favor of examples and short questions.
We believe this strategy is beneficial even in traditional annotation scenarios.

We created separate questionnaires (HITs) for adjectives, adverbs, nouns, and verbs.
A complete example adjective HIT
with directions and questions is shown in Figure \ref{fig:HIT}.
The adverb, noun, and verb questionnaires
had similar questions, but were phrased slightly differently to accommodate differences
in part of speech. These questionnaires are not shown here due to lack of space, but all
four questionnaires are available for download.\footnote{\url{http://www.purl.org/net/saif.mohammad/research}}
The verb questionnaire had an additional question shown in Figure \ref{fig:vquest}.
Since nouns and verbs are not considered gradable, the corresponding questionnaires
did not have Q8 and Q9.
We requested annotations from eight different Turkers for each HIT.


\subsection{The Word Choice Question: Q1}

Q1 is an automatically generated word choice question
that has a clear correct answer. 
It helps identify outlier and malicious annotations. 
If this question is answered incorrectly,
then we assume that the annotator does not know the meanings of the target words,
and we ignore responses to the remaining questions.
Further, as this question makes the annotator think about the meanings of the words
and about the relationship between them, we believe it improves
the responses for subsequent questions.

The options for Q1 were generated automatically.
Each option is a set of four comma-separated words.
The words in the answer are close in meaning
to both of the target words.
In order to create the answer option, we first generated a much larger
source pool of all the words that were in the same thesaurus category as any of the two
target words. (Words in the same category are closely related.)
Words that had the same stem as either of the target words were discarded.
For each of the remaining words,
we added their Lesk similarities with the two target words \cite{Lesk03}.
The four words with the highest sum were chosen to form the answer option.

The three distractor options were randomly selected
from the pool of correct answers for all other word choice questions.
Finally, the answer and distractor options were presented to the
Turkers in random order.\\

\subsection{Post-Processing}

The response to a HIT by a Turker
is called an {\it assignment}.
We obtained about 12,448 assignments in all (1556 pairs $\times$ 8 assignments each).
About 7\% of the adjective, adverb, and noun assignments and
about 13\% of the verb assignments had 
an incorrect answer to Q1. 
These assignments were discarded, leaving
1506 target pairs with three or more valid assignments.
We will refer to this set of assignments as the {\it master set},
and all further analysis in this paper is based on this set.
Table~\ref{tab:assignments} gives a breakdown of the average
number of annotations for each of the target pairs in the master set.

\begin{table}[t]
\centering
\caption{Number of word pairs and average number of annotations per word pair in the master set.}
 {\small
\begin{tabular}{l rr}
\hline
	  part of	&\# of 	&average \# of\\
	  speech	&word pairs	&annotations\\
\hline
	adverbs		&182	&7.80\\
	adjectives	&631	&8.32\\
	nouns		&405	&8.44\\
	verbs		&288	&7.58\\
	{\bf all}	&{\bf 1506}	&{\bf 8.04}\\
\hline
\end{tabular}
 }
\label{tab:assignments}
\end{table}

\subsection{Prevalence of Different Kinds of Contrasting Pairs}

For each question pertaining to every word pair in the master set, we determined the most
frequent response by the annotators. 
Table \ref{tab:responses} gives the percentage of word-pairs in the master set that
received a most frequent response of ``yes''.
The first column in the table lists the question number followed by
a brief description of question. (Note that the Turkers saw only the full forms
of the questions, as shown in the example HIT.)

Observe that most of the word pairs are considered to have at least some contrast in meaning.
This is not surprising since the master set was constructed using words connected through WordNet's antonymy relation.\footnote{All 
of the direct antonyms were marked as contrasting by the Turkers. Only a few indirect antonyms were marked as {\it not contrasting}.}
Responses to Q3 show that not all contrasting pairs are considered opposite,
and this is especially the case for adverb pairs and noun pairs.
The rows in Table~\ref{tab:contra ant} show the percentage of words in the master set
that are contrasting (row 1), opposite (row 2), and contrasting but not opposite (row 3).


Responses to Q5, Q6, Q7, Q8, and Q9 (Table \ref{tab:responses}) show the prevalence of different
kinds of relations and properties of the target pairs. 

\begin{table*}[t]
\caption{Percentage of word pairs that received a response of ``yes'' for the questions in the questionnaire.
`adj.' stands for adjectives. `adv.' stands for adverbs.}
\centering
 {\small
\begin{tabular}{l l rrrr}
\hline
									 & &\multicolumn{4}{c}{\% of word pairs}\\
Question									 &answer &adj.	 &adv. &nouns	&verbs\\
\hline
Q2. Do X and Y have some contrast?  &yes  &99.5	&96.8	&97.6	&99.3\\
Q3. Are X and Y opposites?                    &yes &91.2	&68.6	&65.8	&88.8\\
Q5. Are X and Y at two ends of a dimension?    &yes  &81.8	&73.5	&81.1	&94.4\\
Q6. Does X imply not Y?                        &yes  &98.3	&92.3	&89.4	&97.5\\
Q7. Are X and Y mutually exhaustive?         &yes &85.1	&69.7 	&74.1	&89.5\\
Q8. Does X represent a point on some scale?  &yes &78.5	&77.3  &-		&-\\
Q9. Does Y represent a point on some scale?  &yes &78.5	&70.8  &-		&-\\
Q10. Does X undo Y OR does Y undo X?  	&one way 	&-	&-  &-		&3.8\\
     							  		&both ways  &-	&-  &-		&90.9\\
\hline
\end{tabular}
 }
\label{tab:responses}
\end{table*}

\begin{table*}[]
\caption{Percentage of WordNet source pairs that are contrasting, opposite, and ``contrasting but not opposite''.}
\centering
 {\small
\begin{tabular}{l l rrrr}
\hline
category						&basis				&adj.	&adv. &nouns	&verbs\\
\hline
contrasting						&Q2 yes				&99.5 &96.8 &97.6 &99.3\\
opposites						&Q2 yes and Q3 yes		&91.2 &68.6 &60.2 &88.9\\
contrasting, but not opposite	&Q2 yes and Q3 no		&8.2 &28.2 &37.4 &10.4\\
\hline
\end{tabular}
 }
\label{tab:contra ant}
\end{table*}

%
%

 \begin{table*}[]
 \caption{Percentage of contrasting word pairs belonging to various sub-types. The sub-type ``reversives" applies only to verbs.  The sub-type ``gradable" applies only to adjectives and adverbs.}
 \centering
  {\small
 \begin{tabular}{l l rrrr}
 \hline
 	  sub-type                        &basis              &adv.    &adj. &nouns  &verbs\\
 \hline
 Antipodals						&Q2 yes, Q5 yes		&82.3 &75.9 &82.5 &95.1\\
 Complementaries			  	&Q2 yes, Q7 yes     &85.6 &72.0 &84.8 &98.3\\
 Disjoint			  			&Q2 yes, Q7 no     	&14.4 &28.0 &15.2 &1.7\\
 Gradable 						&Q2 yes, Q8 yes, Q9 yes	&69.6 &66.4 &- &-\\
 Reversives						&Q2 yes, Q10 both ways		&- &- &- &91.6\\

 \hline
 \end{tabular}
  }
 \label{tab:contra kinds}
 \end{table*}

\begin{table*}[]

 \setlength{\baselineskip}{2\baselineskip} 
\caption{Breakdown of answer agreement by target-pair part of speech and question: For every target pair, a question is answered by about eight annotators. 
The majority response is chosen as the answer. The ratio of the size of the
majority and the number of annotators is indicative of the amount of agreement.
The table below shows the average percentage of this ratio.}
\centering
 {\small
\begin{tabular}{l rrrr r}
\hline
question									&adj.	 &adv. &nouns	&verbs &{\bf average}\\
\hline
Q2. Do X and Y have some contrast?   &90.7	&92.1	&92.0	&94.7 &{\bf 92.4}\\
Q3. Are X and Y opposites?                    &79.0	&80.9	&76.4	&75.2 &{\bf 77.9}\\
Q5. Are X and Y at two ends of a dimension?     &70.3	&66.5	&73.0	&78.6 &{\bf 72.1}\\
Q6. Does X imply not Y?                         &89.0	&90.2	&81.8	&88.4 &{\bf 87.4}\\
Q7. Are X and Y mutually exhaustive?         &70.4	&69.2 	&78.2	&88.3 &{\bf 76.5}\\
{\bf average (Q2, Q3, Q5, Q6, and Q7)}		 &{\bf 82.3} &{\bf 79.8}  &{\bf 80.3}	&{\bf 85.0} &{\bf 81.3}\\
Q8. Does X represent a point on some scale?  &77.9	&71.5  &-		&- &{\bf 74.7}\\
Q9. Does Y represent a point on some scale?  &75.2	&72.0  &-		&- &{\bf 73.6}\\
Q10. Does X undo Y OR does Y undo X?       &-  &-  &-      &73.0 &{\bf 73.0}\\ 
\hline
\end{tabular}
 }
\label{tab:agreement}
\vspace*{-4mm}
\end{table*}

Table \ref{tab:contra kinds} shows the percentage of contrasting
word pairs that may be classified into the different types discussed in Section~\ref{sec:kindsofopps} earlier.
Observe that rows for all categories other than the disjoints have percentages greater than 60\%.
This means that a number of contrasting word pairs can be classified into more than one kind.
Complementaries are the most common kind in case of adverbs, nouns, and verbs, whereas antipodals
are most common among adjectives. A majority of the adjective and adverb contrasting pairs
are gradable, but more than 30\% of the pairs are not.
Most of the verb pairs are reversives (91.6\%).
Disjoint pairs are much less common than all the other categories considered, and they are most prominent
among adjectives (28\%), and least among verb pairs (1.7\%).

\subsection{Agreement}

People do not always agree on linguistic classifications of terms,
and one of the goals of this work was to determine how much people agree
on properties relevant to different kinds of opposites.
Table~\ref{tab:agreement} lists the breakdown of agreement by target-pair part of speech and question,
where agreement is the average percentage of the number of Turkers giving the most-frequent response to a question---the
higher the number of Turkers that vote for the majority answer, the higher is the agreement.

Observe that agreement is highest when asked whether a word pair has some degree of
contrast in meaning (Q2), and that there is a marked drop when asked if the two
words are opposites (Q3). This is true for each of the parts of speech, although the drop is highest for verbs (94.7\% to 75.2\%).

For questions 5 through 9, we see varying degrees of agreement---Q6 obtaining the highest
agreement and Q5 the lowest. There is marked difference across parts of speech
for certain questions. For example,
verbs are the easiest to identify (highest agreement for Q5, Q7, and Q8).
For Q6, nouns have markedly lower agreement than all other parts of speech---not surprising considering that
the set of disjoint opposites is traditionally associated with equipollent adjectives
and stative verbs. Adverbs and adjectives have markedly lower agreement scores for Q7 than nouns and verbs.

%
%



\section{Manifestation of highly contrasting word pairs in text}
\label{sec:hypotheses}


As pointed out earlier, there is work on a small
set of opposites showing that opposites co-occur more often than chance \cite{CharlesM89,Fellbaum95}.
Section \ref{sec:cohyp} describes experiments on a larger scale to determine whether
highly contrasting word pairs (including opposites) occur together more often than randomly chosen word pairs
of similar frequency. The section also compares co-occurrence associations
with synonyms. 


Research in distributional similarity has found that entries in distributional thesauri
tend to also contain 
terms that are opposite in meaning  \cite{Lin98B,LinZQZ03}.
Section \ref{sec:disthyp} describes experiments to determine whether highly contrasting word pairs (including opposites) occur in similar contexts 
as often as randomly chosen pairs of words with similar frequencies, and whether
highly contrasting words occur in similar contexts as often as synonyms. 

\subsection{Co-occurrence}
\label{sec:cohyp}
In order to compare the tendencies of highly contrasting word pairs, synonyms, and random word pairs to co-occur in text,
we created three sets of word pairs: the {\it high-contrast set}, 
the {\it synonyms set}, 
and the {\it control set of random word pairs}.
The high-contrast set was created from a pool of direct and indirect opposites (nouns, verbs, and adjectives) from WordNet. 
We discarded pairs that did not meet the following conditions:
(1) both members of the pair must be unigrams,
(2) both members of the pair must occur in the {\it British National Corpus (BNC)} \cite{Burnard00u}, and
(3) at least one member of the pair must have a synonym in WordNet.
A total of 1358 word pairs remained, and these form the high-contrast set.

Each of the pairs in the high-contrast set was used to create a synonym pair
by choosing a WordNet synonym of exactly one member of the pair.\footnote{If both members of a pair have
WordNet synonyms, then one is chosen at random, and its synonym is taken.}
If a word has more than one synonym, then the most frequent synonym is chosen.\footnote{WordNet
lists synonyms in order of decreasing frequency in the SemCor corpus.}
These 1358 word pairs form the synonyms set.
Note that for each of the pairs in the high-contrast set, there is a corresponding pair in the synonyms set,
such that the two pairs have a common term.
For example, the pair {\it agitation} and {\it calmness} in the high-contrast set
has a corresponding pair {\it agitation} and {\it ferment} in the synonyms set.
We will refer to the common terms ({\it agitation} in the above example) as the focus words.
Since we also wanted to compare occurrence statistics of the high-contrast set with the random pairs set,
we created the control set of random pairs by taking each of the focus words and pairing them with another word
in WordNet that has a frequency of occurrence in BNC closest to the term contrasting with the focus word. 
This is to ensure that members of the pairs across the high-contrast set and the control set have similar unigram frequencies.

\begin{table}[t]
 \caption{Pointwise mutual information (PMI) of word pairs. High positive values imply a tendency to co-occur in text more often than random chance.}
 \centering
  {\small
 \begin{tabular}{l  cc}
 \hline
                        &average PMI 	&standard deviation\\
 \hline
	high-contrast set		&1.471			&2.255\\
	random pairs set	&0.032			&0.236\\
	synonyms set		&0.412			&1.110\\	
 \hline
 \end{tabular}
  }
 \label{tab:soa stats}
 \end{table}

We calculated the pointwise mutual information (PMI) \cite{ChurchH89} for each of the word pairs in
the high-contrast set, the random pairs set, and the synonyms set using unigram
and co-occurrence frequencies in the BNC.
If two words occurred within a window of five adjacent words in a sentence, they were
marked as co-occurring (same window as Church and Hanks \shortcite{ChurchH89} used in their
seminal work on word--word associations). 
Table~\ref{tab:soa stats} shows the average and standard deviation in each set.  
Observe that the high-contrast pairs have a much higher tendency to co-occur than the random pairs
control set, and also the synonyms set. However,
the high-contrast set has a large standard deviation. 
A two-sample $t$-test revealed that the high-contrast set is significantly different
from the random set ($p < 0.05$), and also that the high-contrast set is significantly different
from the synonyms set ($p < 0.05$).

However, on average the PMI between a focus word and its contrasting term
was lower than the PMI between the focus word and 
3559 other words in the BNC.
These were often words related to the focus words, but nether contrasting nor synonymous.
Thus, even though a high tendency to co-occur is a feature of highly contrasting pairs, it is
 not a sufficient condition for detecting them.
We use PMI as part of our method for determining the degree of lexical contrast (described ahead in Section \ref{sec:method}).


 \subsection{Distributional similarity}
 \label{sec:disthyp}
 Charles and Miller \shortcite{CharlesM89} proposed that in most contexts, opposites may be
 interchanged.  The meaning of the utterance will be inverted, of course, but the
 sentence will remain grammatical and linguistically plausible.  This came
 to be known as the {\it substitutability hypothesis}.  However, their
 experiments did not support this claim. They found that given a
 sentence with the target adjective removed, most people did not
 confound the missing word with its opposite.  Justeson and Katz \shortcite{JustesonK91} later
 showed that in sentences that contain both members of an adjectival opposite pair, the
 target adjectives do indeed occur in similar syntactic structures at
 the phrasal level.  
 Jones et al.\@ \shortcite{Jones07} show how the tendency to appear in certain textual constructions such as ``from X to Y'' and ``either X or Y'' are indicative of prototypicalness of opposites.
 Thus, 
 we can formulate the {\it distributional hypothesis of highly contrasting pairs}: highly contrasting pairs occur in similar
 contexts more often than non-contrasting word pairs.

 We used the same sets of high-contrast pairs, synonyms, and random pairs described in the previous sub-section 
to gather empirical proof of the
 distributional hypothesis.
 We calculated the
 distributional similarity between each 
 pair in the three sets using
Lin's \shortcite{Lin98B} measure. 
Table~\ref{tab:dsim stats} shows the average and standard deviation in each set.
Observe that the high-contrast set has a much higher average distributional similarity than the random pairs
control set, and interestingly it is also higher than the synonyms set. Once again,
the high-contrast set has a large standard deviation.
A two-sample $t$-test revealed that the high-contrast set is significantly different
from both the random set 
and the synonyms set with a confidence interval of 0.05.
 This demonstrates that relative to other word pairs, high-contrast pairs
 tend to occur in similar contexts.  
We also find that the synonyms set has a significantly higher distributional similarity than the random pairs set ($p < 0.05$).
 This shows that near-synonymous 
  word pairs also occur in similar contexts
 (the distributional hypothesis of similarity).  
Further, a consequence of the large standard deviations in the cases of both high-contrast pairs and synonyms 
means that distributional similarity alone is not sufficient to determine
whether two words are contrasting or synonymous.
An automatic method for recognizing contrast will require additional cues.
Our method 
uses PMI and other sources of information described in the next section.

\begin{table}[t]
 \caption{Distributional similarity of word pairs. The measure proposed in Lin (1998) was used.}
 \centering
  {\small
 \begin{tabular}{l  cc}
 \hline
                        &average distributional similarity 	&standard deviation\\
 \hline
	opposites set		&0.064			&0.071\\
	random pairs set	&0.036			&0.034\\
	synonyms set		&0.056			&0.057\\	
 \hline
 \end{tabular}
  }
 \label{tab:dsim stats}
 \end{table}

\section{Computing Lexical Contrast}
\label{sec:method}
In this section, we recapitulate the automatic method for determining lexical contrast that we first
proposed in Mohammad et al.\@ \shortcite{MohammadDH08}.
Additional details are provided regarding the lexical resources used (Section \ref{sec:resources})
and the method itself (Section \ref{sec:model}).  

\subsection{Lexical Resources}
\label{sec:resources}
Our method makes use of a published thesaurus and co-occurrence information from text. Optionally,
it can use opposites listed in WordNet if available. We briefly describe these resources here.
\subsubsection{Published thesauri}
Published thesauri, such as {\it Roget's} and {\it Macquarie},
divide the vocabulary of a language into about a thousand {\it categories}. 
 Words within a category are semantically related to each other, and they tend to pertain to a coarse concept. 
Each category is represented by a category number (unique ID) and a {\it head word} ---
a word that best represents the meanings of the words in the category.
One may also find opposites in the same
category, but this is rare. 
Words with more than one meaning
may be found in more than one category; these represent its coarse
senses.

Within a category, the words are grouped into finer units called {\it paragraphs}. Words in the
same paragraph are closer in meaning than those in differing
paragraphs. Each paragraph has a {\it paragraph head} --- a word that best represents
the meaning of the words in the paragraph.
Words in a thesaurus paragraph belong to the same part of speech.
A thesaurus category may have multiple paragraphs belonging to the same part of speech.
For example, a category may have three noun paragraphs, four verb paragraphs, and one adjective paragraph.
We will take advantage of the structure of the thesaurus in our approach.


\subsubsection{WordNet}
As mentioned earlier, WordNet encodes certain opposites.
However, we found in our experiments (Section \ref{sec:eval} below) that more than 90\% of contrasting pairs
included in Graduate Record Examination (GRE) ``most contrasting word'' questions are not encoded in WordNet.
Also, neither WordNet nor any other manually-created repository of opposites provides the {\it degree} of contrast between word pairs.
Nevertheless, we investigate the usefulness of WordNet
as a source of seed opposites for our approach.

\subsection{Proposed Measure of Lexical Contrast}
\label{sec:model}
Our method for determining lexical contrast has two parts: (1) determining whether the target word pair
is contrasting or not, and (2) determining the degree of contrast between the words.
\subsubsection{Detecting whether a target word pair is contrasting}
\label{sec:concats}

We use the contrast hypothesis to determine whether two words are contrasting. The hypothesis is repeated below:
\vspace*{-6mm}
 \begin{quote}
{\it Contrast Hypothesis}: 
If a pair of words, $A$ and $B$, are contrasting, then there is a pair of opposites, $C$ and $D$, such that $A$ and $C$ are strongly related and $B$ and $D$ are strongly related.
\end{quote}
\noindent Even if a few exceptions to this hypothesis are found (we are not aware of any), the hypothesis would remain useful for practical applications.
We first determine pairs of thesaurus categories that have at least one word in each category that are opposites of each other. 
We will refer to these categories as {\it contrasting categories} and the opposite connecting the two categories as the {\it seed opposite}.
Since each thesaurus category is a collection of closely related terms, all of the word pairs across two contrasting categories satisfy the contrast hypothesis,
and they are considered to be contrasting word pairs.
Note also that words within a thesaurus category may belong to different parts of speech, and they may be related
to the seed opposite word through any of the many possible semantic relations. Thus a small number of seed opposites
can help identify a large number of contrasting word pairs.

We determine whether  two categories are contrasting using the three methods described below,
which may be used alone or in combination with each other:\\



\noindent {\bf Method 1: Using word pairs generated from affix patterns.}\\
Opposites such as {\it hot--cold} and {\it dark--light} occur
frequently in text, but in terms of type-pairs they are outnumbered by
those created using affixes, such as {\it un-} ({\it clear--unclear})
and {\it dis-} ({\it honest--dishonest}).  Further, this phenomenon
is observed in most languages \cite{Lyons77}.

\begin{table*}
    \caption{Fifteen aff\/ix patterns used to generate opposites. Here `X' stands for any sequence of letters common to both words $w_1$ and $w_2$.}
    \centering
{\small
    \begin{tabular}{rrl r l}
    \hline
	& \multicolumn{2}{c}{\bf affix pattern}	& &\\
{\bf pattern \#}	&{\bf word 1}	&{\bf word 2} 	&{\bf \# word pairs}	&{\bf example pair}\\
 \hline
1	& X 	&{\it anti}X 			&41	&{\it clockwise--anticlockwise}\\
2	& X  &{\it dis}X 			&379	&{\it interest--disinterest}\\	
3	& X  &{\it im}X 				&193	&{\it possible--impossible}\\	
4	& X  &{\it in}X          	&690	&{\it consistent--inconsistent}\\
5	& X  &{\it mal}X            	&25	&{\it adroit--maladroit}\\	
6	& X  &{\it mis}X             	&142	&{\it fortune--misfortune}\\ 
7	& X  &{\it non}X             	&72	&{\it aligned--nonaligned}\\
8	& X  &{\it un}X          		&833	&{\it biased--unbiased}\\
9	& {\it l}X &{\it ill}X        &25	&{\it legal--illegal}\\
10	& {\it r}X &{\it irr}X         &48	&{\it regular--irregular}\\
11	& {\it im}X &{\it ex}X           &35	&{\it implicit--explicit}\\
12	& {\it in}X &{\it ex}X           &74	&{\it introvert--extrovert}\\
13	& {\it up}X &{\it down}X         &22	&{\it uphill--downhill}\\
14	& {\it over}X &{\it under}X      &52	&{\it overdone--underdone}\\
15	& X{\it less} 	&X{\it ful}	   &51	&{\it harmless--harmful}\\
   \hline
	&\multicolumn{2}{c}{Total:}	&2682 &\\
   \hline
    \end{tabular}
	}
    \label{tab:rules}
\end{table*}


Table~\ref{tab:rules} lists fifteen  affix patterns that tend to
generate opposites in English.  They were compiled by the first author
by examining a small list of affixes for the English language.\footnote{\url{http://www.englishclub.com/vocabulary/prefixes.htm}}
These patterns were applied to all
words in the thesaurus that are at least three characters long.
If the resulting term was
also a valid word in the thesaurus, then the word-pair was added to the
{\it affix-generated seed set}.  These fifteen rules generated 2,682
word pairs when applied to the words in the {\it Macquarie Thesaurus}.
Category pairs that had these opposites were marked as contrasting.
Of course, not all of the word pairs generated through affixes are
truly opposites, for example {\it sect--insect} and {\it part--impart}.
For now, such pairs are sources of error in the system.
Manual analysis of these 2,682 word pairs can help determine whether
this error is large or small. 
(We have released the full set of word pairs.)
However, evaluation results (Section 7) indicate that
these seed pairs improve the overall accuracy of the system.

\begin{figure}[t]
\begin{center}
\includegraphics[width=0.8\columnwidth]{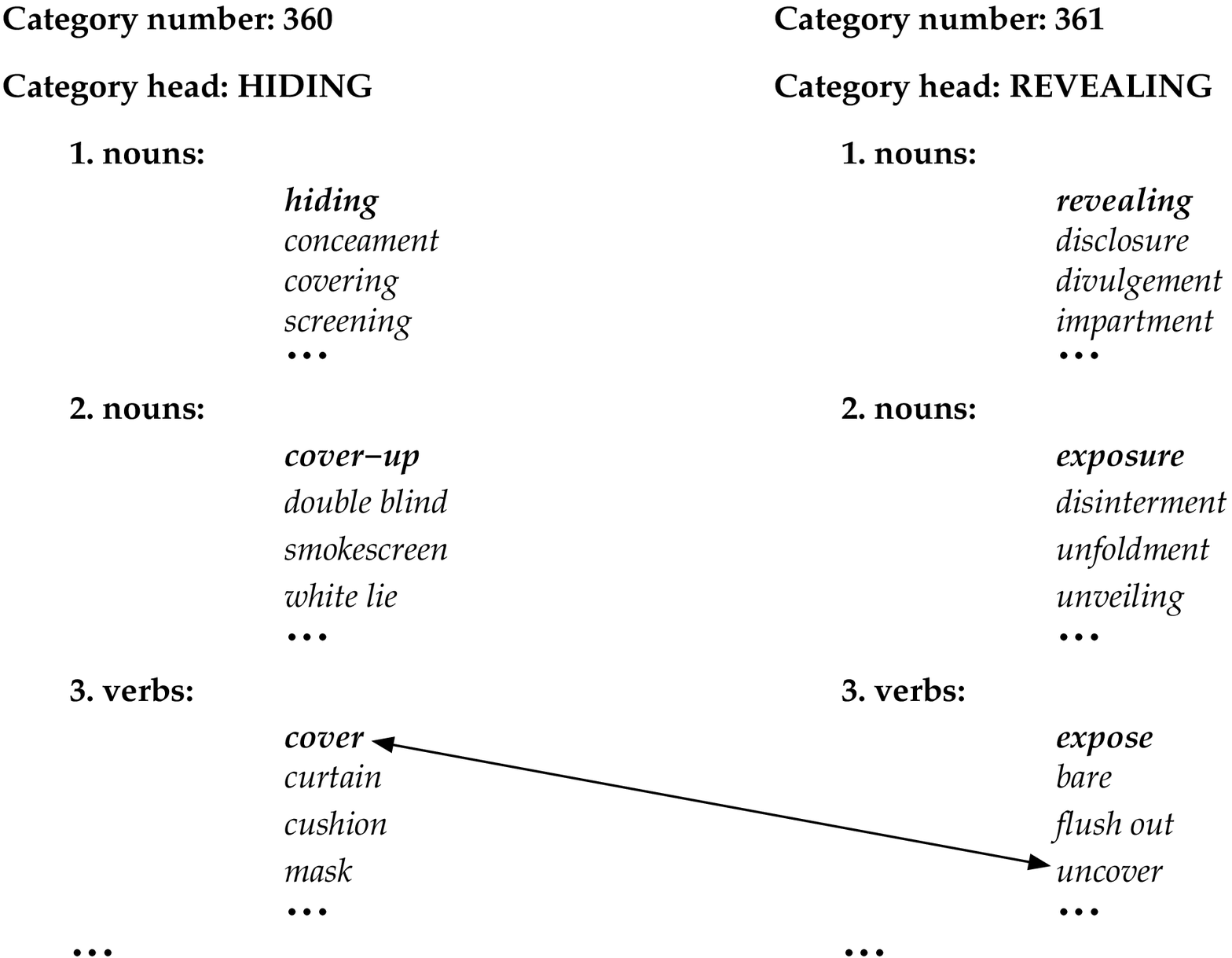}
\end{center}
\caption{Example contrasting category pair. The system identifies the pair to be contrasting through the affix-based
seed pair {\it cover--uncover}. The paragraphs of {\bf cover} and {\bf expose} are referred to as prime contrasting paragraphs.
Paragraph heads are shown in bold italic.}
\label{fig:oppCats}
\end{figure}

Figure \ref{fig:oppCats} presents such an example pair. Observe that
categories 360 and 361 have the words {\it cover} and {\it uncover}, respectively.
Affix pattern 8 from Table 1 produces seed pair {\it cover--uncover},
and so the system concludes that the two categories have contrasting meaning.
 The contrast in meaning is especially strong for the paragraphs {\bf \it cover} and {\bf \it expose}
 because words within these paragraphs are very close in meaning to {\it cover} and {\it uncover},
 respectively. We will refer to such thesaurus paragraph pairs that have one word each of a seed pair
 as {\it prime contrasting paragraphs}. We expect the words across prime contrasting
 paragraphs to have a high degree of antonymy (for example, {\it mask} and {\it bare}), whereas words across other contrasting category paragraphs
 may have a smaller degree of antonymy as the meaning of these words may diverge significantly from the meanings
 of the words in the prime contrasting paragraphs (for example, {\it white lie} and {\it disclosure}).\\



\noindent {\bf Method 2: Using opposites from WordNet.}\\
We compiled a list of 20,611 pairs that WordNet records as direct and indirect opposites.
(Recall discussion in Section 4 about direct and indirect opposites.)
A large number of these pairs include multiword expressions. Only 10,807 of the
20,611 pairs have both words in the {\it Macquarie
Thesaurus}---the vocabulary used for our experiments.  We will refer
to them as the {\it WordNet seed set}.
Category pairs that had these opposites were marked as contrasting.\\

\noindent {\bf Method 3: Using word pairs in adjacent thesaurus categories.}\\
Most published thesauri, such as {\it Roget's}, are organized such that categories corresponding to opposing concepts
are placed adjacent to each other. For example, in the {\it Macquarie Thesaurus}: category 369 is about honesty
and category 370 is about dishonesty; as shown in Figure \ref{fig:oppCats}, category 360 is about
hiding and category 361 is about revealing. There are a number of exceptions to this rule, and often a
category may be contrasting in meaning to several other categories.
However, since this was an easy-enough heuristic to implement, we
investigated the usefulness of considering adjacent thesaurus categories as contrasting.
We will refer to this as the {\it adjacency heuristic}.
 Note that this method of determining contrasting categories does not explicitly
identify a seed opposite, but one can assume the head words of these category pairs as the seed opposites.

To determine how accurate the adjacency heuristic is, the first author
manually inspected adjacent thesaurus categories in the {\it Macquarie Thesaurus} to determine which of them
were indeed contrasting. Since a category, on average, has about a hundred words, 
the task was made less arduous by representing each category by just the first ten words listed in it.
This way it took only about five hours to manually determine that 209 pairs of the 811 adjacent Macquarie category pairs
were contrasting.
Twice, it was found that category number X was contrasting not just to category number X+1 but also to category number
X+2: category 40 ({\sc aristocracy}) has a meaning that contrasts that of category 41 ({\sc middle class}) as well as category 42 ({\sc working class});
category 542 ({\sc past}) contrasts with category 543 ({\sc present}) as well as category 544 ({\sc future}).
Both these X--(X+2) pairs are also added to the list of manually annotated contrasting categories.

\subsubsection{Computing the degree of contrast between two words}
\label{sec:degant}

$\,$ Charles and Miller \shortcite{CharlesM89} and Fellbaum \shortcite{Fellbaum95} argued that opposites tend to co-occur more often than random chance.
Murphy and Andrew~\shortcite{MurphyA93} claimed that
the greater-than-chance co-occurrence of opposites is because together they
convey contrast well, which is rhetorically useful.
 We showed earlier in Section 5.1 that highly contrasting pairs (including opposites) co-occur more often than randomly chosen pairs.
 All of these support the degree of contrast hypothesis stated earlier in the introduction:
\begin{quote}
{\it Degree of Contrast Hypothesis}: If a pair of words, $A$ and $B$, are contrasting, then their
degree of contrast is proportional to their tendency to co-occur in a large corpus.
\end{quote}


We used PMI to
capture the tendency of word--word co-occurrence.
We collected these co-occurrence statistics from 
the {\it Google n-gram corpus} \cite{BrantsF06},
which was created from a text collection of over 1 trillion words.
 Words that occurred within a window of 5 words were considered to be
 co-occurring.

%



\begin{figure}[t]
\begin{center}
\includegraphics[width=0.8\columnwidth]{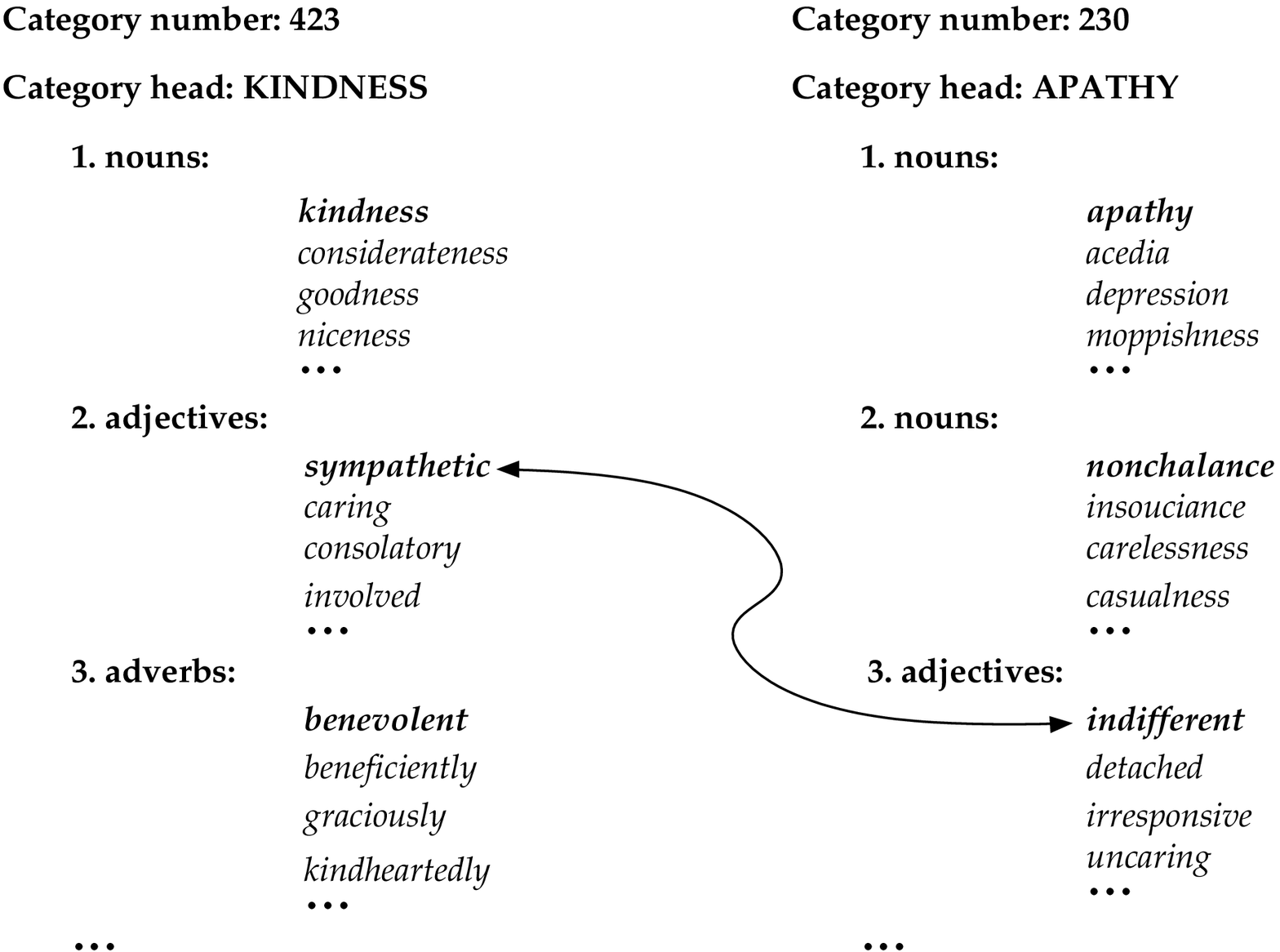}
\end{center}
\caption{Example contrasting category pair that has Class II and Class III contrasting pairs. The system identifies the pair to be contrasting through the affix-based
seed pair {\it caring} (second word in paragraph 2 or category 423) and {\it uncaring} (fourth word in paragraph 3 or category 230). The paragraphs of {\it sympathetic} and {\it indifferent} are therefore the prime contrasting paragraphs and so all word pairs
that have one word each from these two paragraphs are Class II contrasting pairs. All other pairs formed by taking one word each from the two contrasting categories are the Class III contrasting pairs.
Paragraph heads are shown in bold italic.}
\vspace*{-4mm}
\label{fig:oppCats2}
\end{figure}

We expected that some features may be more accurate than others. If multiple features give evidence towards
opposing information, then it is useful for the system to know which feature is more reliable.
Therefore, we held out some data from the evaluation data described in Section 7.1 as the
development set. Experiments on the development set showed that 
contrasting words may be placed in
three bins corresponding to the amount of reliability of the source feature: high, medium, or acceptable.
\begin{itemize}
\item {\bf High reliability (Class I):}  target words that belong to adjacent thesaurus categories. %
For example, all the word pairs across categories 360 and 361, shown in Figure 3. 
Examples of class I contrasting word pairs from the development set include
{\it graceful--ungainly, fortunate--hapless, obese--slim,} and {\it effeminate--virile}.
(Note, there need not be any affix or WordNet seed pairs across adjacent thesaurus categories for these word pairs to be marked Class I.)
As expected, if we use only those adjacent categories that were manually identified to be contrasting (as described in Section 6.2.1. method 3),
then the system obtains even better results than those obtained using all adjacent thesaurus categories. (Experiments and results
shown ahead in Section 7.1).
\item {\bf Medium reliability (Class II):} target words that are not Class I contrasting pairs, but belong to one paragraph each of a prime contrasting paragraph.
For example, all the word pairs across the paragraphs of {\it sympathetic} and {\it indifferent}. See Figure 4.
Examples of class II contrasting word pairs from the development set include
{\it altruism--avarice, miserly--munificent, accept--repudiate,} and {\it improper--prim}.
\item {\bf Acceptable reliability (Class III):} target words that are not Class I or Class II contrasting pairs,
but occur across contrasting category pairs. 
For example, all word pairs across categories 423 and 230 except those that have one word each from the paragraphs of {\it sympathetic} and {\it indifferent}. See Figure 4.
Examples of class III contrasting word pairs from the development set include
{\it pandemonium--calm, probity--error, artifice--sincerity,} and {\it hapless--wealthy}.
\end{itemize}

Even with access to very large textual datasets, there is always a long tail of words
that occur so few times that there is not enough co-occurrence information for them.
Thus we assume that all word pairs in
Class I have a higher degree of contrast than all word pairs in Class II, and that all word pairs in Class II 
have a higher degree of contrast than the pairs in Class III. If two word pairs belong to the same class, then we calculate
their tendency to co-occur with each other in text to determine which pair is more contrasting.
All experiments in the evaluation section ahead follow this method.

%
\subsubsection{Lexicon of contrasting word pairs}
Using the method described in the previous sub-sections, we generated
a lexicon of word pairs pertaining to Class I 
and Class II. 
The lexicon has 6.3 million contrasting word pairs, about 3.5 million of which belong to Class I and 
about 2.8 million to Class II.
Class III pairs are even more numerous and given a word pair, our algorithm 
checked whether it is a class III pair, but we did not create a complete set of all Class III contrasting pairs.
 Class I and II lexicons are available for download and summarized in Table~\ref{tab:data}.


\section{Evaluation}
\label{sec:eval}
We evaluate our algorithm on two different tasks and four datasets. Section 7.1 describes experiments on solving existing GRE ``choose the most contrasting word'' questions
(a recapitulation of the evaluation reported in Mohammad et al.\@ \shortcite{MohammadDH08}).
Section 7.2 describes experiments on solving newly created ``choose the most contrasting word'' questions specifically designed to determine performance on different kinds of opposites.
And lastly, Section 7.3 describes experiments on two different datasets where the goal is to identify whether a given word pair is synonymous or antonymous.



\subsection{Solving GRE's ``choose the most contrasting word'' questions}
\label{sec:gre}
 The Graduate Record Examination (GRE) is a test taken by thousands of North American graduate school applicants. The test is administered by Educational Testing Service (ETS).
The Verbal Reasoning section of GRE is designed to test verbal skills.
Until August 2011, one of its sub-sections had a set of questions pertaining to word-pair contrast. Each question had a 
target word and four or five alternatives, or option words.
The objective was to identify the alternative which was most contrasting with respect to the
target. For example, consider:
\begin{quote}
 \begin{tabular}{l l l l l l }
 {\bf \em adulterate:}	
 &a. {\it renounce}	&b. {\it forbid}	
 &c. {\it purify}	&d. {\it criticize}	&e. {\it correct}\\
 \end{tabular}
\end{quote}
\noindent Here the target word is {\it adulterate}. One of the alternatives provided
is {\it correct}, which as a verb has a meaning that contrasts with that of {\it
adulterate}; however, {\it purify} has a greater degree of contrast with
{\it adulterate} than {\it correct} does and must be chosen
in order for the instance to be marked as correctly answered.
 ETS referred to these questions as ``antonym questions'', where the examinees had to ``choose the word
most nearly opposite'' to the target. However, most of the target--answer pairs
are not gradable adjectives, and since most of them are not opposites either, we will refer
to these questions as ``{\it choose the most contrasting word'' questions} or {\it contrast questions} for short.

Evaluation on this dataset tests whether the automatic method is able to identify
not just opposites but also those pairs that are not opposites but that have some degree of semantic contrast. Notably,
for these questions, the method must be able to identify that one word pair has
a higher degree of contrast than all others, even though that word pair may not necessarily be an opposite.

\subsubsection{Data}
A web search for large sets of contrast questions
yielded two independent sets of questions designed to prepare students for the
GRE. 
The first set consists of 162 questions. We used this set while we were developing our lexical contrast algorithm 
described in Section~\ref{sec:crowd}.
Therefore, will refer to it as the {\it development set}. 
The development set helped determine which features of lexical contrast were more reliable than others.
The second set has 1208 contrast questions. 
We discarded questions that had a multiword target or
alternative. After removing duplicates we
were left with 790 questions,
which we used as the unseen {\it test set}.
This dataset was used (and seen) only after our algorithm for determining lexical contrast was frozen.


Interestingly, the data contains many instances that have the same
target word used in different senses. For example:

\begin{quote}
 \begin{tabular}{l l l l l l}
 1. {\bf \em obdurate:}	
 &a. {\it meager}	&b. {\it unsusceptible}	&c. {\it right}		&d. {\it tender}		&e. {\it intelligent}\\
 2. {\bf \em obdurate:}
 &a. {\it yielding}	&b. {\it motivated}	&c. {\it moribund}	&d. {\it azure}		&e. {\it hard}\\
 3. {\bf \em obdurate:}
 &a. {\it transitory}	&b. {\it commensurate} &c. {\it complaisant}	&d. {\it similar}		&e. {\it laconic}\\
 \end{tabular}
\end{quote}
%
%
\noindent In (1), {\it obdurate} is used in the sense of {\sc
hardened in feelings}
and is most contrasting with {\it tender}. In (2),
it is used in the sense of {\sc resistant to persuasion} and is most contrasting with
 {\it yielding}.
In (3),
it is used in the sense of {\sc persistent} and is most contrasting with
 {\it transitory}. 

The datasets also contain questions in which one or more of the alternatives
is a near-synonym of the target word. For example:

\begin{quote}
 \begin{tabular}{l l l l l l}
 {\bf \em astute:}	
 &a. {\it shrewd}	&b. {\it foolish}	&c. {\it callow}		&d. {\it winning}		&e. {\it debating}\\
 \end{tabular}
\end{quote}
\noindent Observe that {\it shrewd} is a near-synonym of {\it astute}. The word most contrasting with
of {\it astute} is {\it foolish}. A manual check of a randomly selected set of 100 test-set questions 
revealed that, on average, one in four had a near-synonym as one of the alternatives.

\subsubsection{Results}
Table \ref{tab:results} presents results obtained on the development and test data
using two baselines, a re-implementation of the method described in Lin et al.\@ (2003), and variations of our method. 
Some of the results are for systems that refrain from attempting questions for which they do not
have sufficient information. We therefore report precision (P),
recall (R), and balanced F-score (F).
\begin{eqnarray}
P &=& \frac{\textrm {\# of questions answered correctly}}{\textrm {\# of questions attempted}}\\
R &=& \frac{\textrm {\# of questions answered correctly}}{\textrm {\# of questions}}\\
F &=& \frac{2 \times P \times R}{P + R}
\end{eqnarray}
\vspace*{-7mm}

\begin{table*}
    \caption{Results obtained on contrast questions. The best performing system and configuration are shown in bold.}
    \centering
 	\resizebox{\textwidth}{!}{
    \begin{tabular}{lcccccccc}
    \hline
	 								&\multicolumn{3}{c}{\bf development data} & &  &\multicolumn{3}{c}{\bf test data}\\
						&P &R &F & & 	&P &R &F\\
	\hline 
	{\it Baselines:}	&   &   &   & & &   &   &\\
	$\;\;\;$ a. random baseline					 	&0.20 &0.20 &0.20 & & 			&0.20 &0.20 &0.20\\
	$\;\;\;$ b. WordNet antonyms	&0.23   &0.23   &0.23   & & &0.23   &0.23   &0.23\\
	{\it Related work:}		&   &   &   & & &   &   &\\
	$\;\;\;$ a. Lin et al.\@ (2003) &0.23   &0.23   &0.23   & & &0.24   &0.24   &0.24\\
	{\bf \emph{Our method:}}		&   &   &   & & &   &   &\\
	$\;\;\;$ a. affix-generated pairs as seeds  		 	&0.72 &0.53 &0.61 & & 			&0.71 &0.51 &0.59\\
	$\;\;\;$ b. WordNet antonyms as seeds 				&0.79 &0.52 &0.63 & &			&0.72 &0.49 &0.58\\
	$\;\;\;$ c. both seed sets (a + b)	 				&0.77 &0.65 &0.70 & & 			&0.72 &0.58 &0.64\\
	$\;\;\;$ d. adjacency heuristic only					&0.81 &0.43 &0.56 & & 		&0.83 &0.44 &0.57\\
	$\;\;\;$ e. manual annotation of adjacent categories			&0.88	&0.41	&0.56	& &		&0.87 	&0.41 	&0.55\\
	$\;\;\;$ f. affix seed set and adjacency heuristic (a + d)		&0.75 &0.60 &0.67 & & 			&0.76 &0.60 &0.67\\
	$\;\;\;$ g. both seed sets and adjacency heuristic (a + b + d) 		&0.76 &0.66 &0.70 & & 	&0.76 &0.63 &0.69\\
	$\;\;\;$ h. affix seed set and annotation of adjacent			&0.79	&0.63	&0.70	& &	&0.78 	&0.60 	&0.68\\
	 $\;\;\;\;\;\;$ categories (a + e)	&	&	&	& &	& 	& 	&\\
	$\;\;\;$ i. {\bf both seed sets and annotation of adjacent} 		&{\bf 0.79}	&{\bf 0.66}	&{\bf 0.72}	& &	&{\bf 0.77} 	&{\bf 0.63} 	&{\bf 0.69}\\
	 $\;\;\;\;\;\;$ {\bf categories (a + b + e)}	&	&	&	& &	& 	& 	&\\
	\hline
    \end{tabular}
	}
	\label{tab:results}
 	\normalsize
\end{table*}

\paragraph{Baselines}

If a system randomly guesses one of the five alternatives with equal probability ({\it random baseline}), then
it obtains an accuracy of 0.2.
A system that looks up the list of WordNet antonyms (10,807 pairs) to solve the contrast
questions is our second baseline. 
However, that obtained the correct answer in only 5 instances of the development set (3.09\% of the 162 instances)
and 25 instances of the test set (3.17\% of the 790 instances). Even if the system guesses at random for all
other instances, it attains only a modest improvement over the random baseline (see row $b$, under ``{\it Baselines}'', in Table~\ref{tab:results}).

\paragraph{Re-implementation of related work}

In order to estimate how well the method of Lin et al.\@ (2003) performs on this task, we
re-implemented their method. For each closest-antonym question, we determined frequency counts 
in the Google n-gram corpus for the phrases 
``from $\langle$target word$\rangle$ to $\langle$known correct answer$\rangle$'',
``from $\langle$known correct answer$\rangle$ to $\langle$target word$\rangle$'',
``either $\langle$target word$\rangle$ or $\langle$known correct answer$\rangle$'', and
``either $\langle$known correct answer$\rangle$ or $\langle$target word$\rangle$''.
We then summed up the four counts for each contrast question. This resulted in non-zero counts
for only 5 of the 162 instances in the development set (3.09\%), and 35 of the 790 instances in the test set (4.43\%).
Thus, these patterns fail to cover a vast majority of closest-antonyms, and even if the system
guesses at random for all other instances, it attains only a modest improvement over the baseline
(see row $a$, under ``Related work'', in Table~\ref{tab:results}).

\paragraph{Our method}


Table \ref{tab:results} presents results obtained on the development and test data
using different combinations of the seed sets and the adjacency heuristic. 
The best performing system is marked in bold.
It has significantly higher precision and recall than that of the method proposed by Lin et al.\@ \shortcite{LinZQZ03}, with 95\% confidence according to the Fisher Exact Test \cite{Agresti90}.

We performed experiments on the development set first, using our method with
configurations described in rows a, b, and d. These results showed that marking
adjacent categories as contrasting has the highest precision (0.81), followed by
using WordNet seeds (0.79), followed by the use of affix rules to generate seeds (0.72).
This allowed us to determine the relative reliability of the three features as described in Section 6.2.2 earlier.
We then froze all system development and ran the remaining experiments, including those on the test data.

Observe that all of the results shown in Table \ref{tab:results} are well above the random baseline of 0.20.
Using only the small set of 
fifteen affix rules, the system performs almost as well as when it uses 10,807 WordNet opposites.
Using both the affix-generated and the WordNet seed sets, the system obtains markedly improved precision and coverage. 
Using only the adjacency heuristic gave precision values (upwards of 0.8) with substantial coverage (attempting more than half of the questions).
Using the manually identified contrasting adjacent thesaurus categories gave precision values just short of 0.9.
The best results were obtained using both seed sets and the contrasting adjacent thesaurus categories (F-scores of 0.72
and 0.69 on the development and test set, respectively).

In order to determine whether our method works well with thesauri other than the {\it Macquarie Thesaurus},
we determined performance of configurations a, b, c, d, f, and h using the 1911 US edition of the {\it Roget's Thesaurus},
which is available freely in the public domain.\footnote{\url{http://www.gutenberg.org/ebooks/10681}}
The results were similar to those obtained using 
the {\it Macquarie Thesaurus}. For example, configuration g obtained a precision of 0.81, recall of 0.58,
and F-score of 0.68 on the test set. 
It may be possible to obtain even better results by combining multiple lexical resources; however, that is left for future work. 
The remainder of this paper reports results obtained with the {\it Macquarie Thesaurus};  the 1911 vocabulary is less suited for practical use in the twenty-first century.

\subsubsection{Discussion}
These results show that our method performs well on questions designed to be challenging for humans.
In tasks that require
higher precision, using only the contrasting adjacent categories is best, whereas
in tasks that require both precision and coverage, the seed sets may
be included. 
Even when both seed sets were included, only four instances in the
development set and twenty in the test set had target--answer pairs
that matched a seed opposite pair. For all remaining instances,
the approach had to generalize to determine the most contrasting word.
This also shows that even the seemingly large number
of direct and indirect antonyms from WordNet (more than 10,000) are by themselves insufficient.


The comparable performance obtained using the affix rules alone 
suggests that even in languages that do not have a WordNet-like resource, substantial
accuracies may be obtained. Of course, improved results when using WordNet antonyms as well suggests that the
information they provide is complementary.

Error analysis revealed that at times the system failed to
identify that a category pertaining to the target word contrasted with a
category pertaining to the answer. 
Additional methods to identify seed opposite pairs will help in such cases.
Certain other errors occurred because one or more alternatives other than
the official answer were also contrasting with the target. For
example, one of the questions has {\it chasten} as the target word. One of the alternatives
is {\it accept}, which has some degree of contrast in meaning to the target. However,
another alternative, {\it reward}, has an even higher degree of contrast
with the target. In this instance, the system erred by choosing {\it accept}
as the answer.


\subsection{Determining performance of automatic method on different kinds of opposites}
The previous subsection showed the overall performance of our method. However, the performance of a method may vary
significantly on different subsets of data.
In order to determine performance on 
 different kinds of opposites,
  we generated new contrast questions from the crowdsourced term pairs 
described earlier in Section~\ref{sec:crowd}.
 Note that for solving contrast questions with this dataset, again the method must be able to identify that one word pair has
a higher degree of contrast than the other pairs; however, unlike the previous sub-section,
here the correct answer is often an opposite of the target.

\subsubsection{Generating Contrast Questions}

For each word pair from the list of WordNet opposites, we chose one word randomly to be the target word,
and the other as one of its candidate options. Four other candidate options
were chosen from Dekang Lin's distributional thesaurus \cite{Lin98B}.\footnote{\url{http://webdocs.cs.ualberta.ca/~lindek/downloads.htm}} 
An entry in the distributional thesaurus has a focus word and a number of other
words that are distributionally similar to the focus word. The words are listed in
decreasing order of similarity. Note that these entries include not just near-synonymous 
words but also at times contrasting words because contrasting words tend to be distributionally similar \cite{LinZQZ03}. 

For each of the target words in our contrast questions, we chose the four 
distributionally closest words from Lin's thesaurus to be the distractors.
If a distractor had the same first three letters as the target word or the correct answer,
then it was replaced with another word from the distributional thesaurus.
This ad-hoc filtering criterion is effective at discarding distractors that
are morphological variants of the target or the answer. For example,
if the target word is {\it adulterate}, then words such as {\it adulterated}
and {\it adulterates} will not be included as distractors even if they
are listed as closely similar terms in the distributional thesaurus.

We place the four distractors and the correct answer in random order.
Some of the WordNet opposites were not listed in Lin's thesaurus,
and the corresponding question was not generated.
In all, 1269 questions were generated.
We created subsets of these questions corresponding to the different kinds of
opposites and also corresponding to different parts of speech.
Since a word pair may be classified as more than one kind of opposite,
the corresponding question may be part of more than one subset.

\subsubsection{Experiments and Results}

We applied our method of lexical contrast to solve
the complete set of 1269 questions and also the various subsets.
Since this test set is created from WordNet opposites, we applied the 
algorithm without the use of WordNet seeds (no WordNet information was used by the method).

 \begin{table}[t]
 \caption{Percentage of contrast questions correctly answered by the automatic method,
 where different questions sets correspond to target--answer pairs of {\it different kinds}.
 The automatic method did not use WordNet seeds for this task. The results shown for `ALL'
 are micro-averages, that is, they are the results for the master set of 1269 contrast questions.}
 \centering
  {\small
 \begin{tabular}{l rrrr}
 \hline
                                        &\# instances &P    &R &F\\
 \hline
 Antipodals                              &1044 &0.95   &0.84   &0.89   \\
 Complementaries                         &1042 &0.95   &0.83   &0.89   \\
 Disjoint                                &228 &0.81   &0.59   &0.69   \\
 Gradable                                &488 &0.95   &0.85   &0.90   \\
 Reversives                              &203 &0.93	 &0.74	 &0.82   \\
 ALL									 &1269 &0.93   &0.79  &0.85   \\
 \hline
 \end{tabular}
  }
\label{tab:acc on kinds}
  \vspace*{-2mm}
 \end{table}

Table~\ref{tab:acc on kinds} shows the precision (P), recall (R),
and F-score (F) obtained by the method on the datasets corresponding
to different kinds of opposites.
The column `\# instances' shows the number of questions in each of the datasets.
The performance of our method on the complete dataset
is shown in the last row ALL. Observe that the F-score of 0.85
is markedly higher than the score obtained on the GRE-preparatory questions.
This is expected because the GRE questions
involved vocabulary from a higher reading level, and included
carefully chosen distractors to confuse the examinee. 
The automatic method obtains highest F-score on the datasets of gradable adjectives (0.90), 
antipodals (0.89), and complementaries (0.89). The precisions and recalls
for these opposites are significantly higher than those of
disjoint opposites. The recall for reversives is also significantly
lower than that the gradable adjectives, antipodals, and complementaries,
but precision on reversives is quite good (0.93).

Table~\ref{tab:acc on pos} shows the precision, recall,
and F-score obtained by the method on the the datasets corresponding
to different parts of speech.
Observe that performances on all parts of speech are fairly high. The method deals with adverb pairs best (F-score of 0.89), 
and the lowest performance is for verbs (F-score of 0.80).
The precision values obtained between on the data from any two parts of speech are not significantly different.
However, the recall obtained on the adverbs is significantly higher than that obtained on adjectives,
and the recall on adjectives is significantly higher than that obtained on verbs.
The difference between the recalls on adverbs and nouns is not significant.
We used the Fisher Exact Test and a confidence interval of 95\% for all significance testing 
reported in this section.

 \begin{table}[t]
 \caption{Percentage of contrast questions correctly answered by the automatic method, where different questions sets correspond to {\it different parts-of-speech}.}
 \centering
  {\small
 \begin{tabular}{l  rrrr}
 \hline
                                        &\# instances &P    &R &F\\
 \hline
 Adjectives                              &551 &0.92   &0.79   &0.85\\
 Adverbs                                 &165 &0.95   &0.84   &0.89\\
 Nouns	                                 &330 &0.93   &0.81   &0.87\\
 Verbs		 				             &226 &0.93   &0.71   &0.80\\
 ALL									 &1269 &0.93   &0.79  &0.85   \\
 \hline
 \end{tabular}
  }
 \vspace*{-4mm}
 \label{tab:acc on pos}
 \end{table}



\subsection{Distinguishing synonyms from opposites}
Our third evaluation follows that of Lin et al.\@ \shortcite{LinZQZ03} and Turney \shortcite{Turney08}.
We developed a system for automatically distinguishing synonyms from opposites, and applied it to 
two datasets. The approach and experiments are described below.

\subsubsection{Data}
Lin et al.\@ \shortcite{LinZQZ03} compiled 80 pairs of synonyms
and 80 pairs of opposites from the Webster's Collegiate Thesaurus \cite{Kay88}
such that each word in a pair is also in their list of the 50 distributionally most similar
words of the other. (Distributional similarity was calculated using the algorithm
proposed by Lin et al.\@ \shortcite{Lin98B}.) Turney \shortcite{Turney08} compiled 136 pairs of words
(89 opposites and 47 synonyms) 
from various websites
for learners of English as a second language (ESL);
the objective for the learners is to identify whether the words in a pair are opposites or synonyms
of each other.
 The goals of this evaluation are to determine whether our automatic method can distinguish opposites
from near-synonyms, and to compare our method with the closest related work on an evaluation task for which published results are already available.


\subsubsection{Method}
\label{sec:ASmethod}
The core of our method is this:
\vspace*{-3mm}
\begin{enumerate}
\item Word pairs that occur in the same thesaurus category are close in meaning and so are marked as synonyms.
\vspace*{-1mm}
\item Word pairs that occur in contrasting thesaurus categories or paragraphs (as described in Section \ref{sec:concats} above)
are marked as opposites.
\end{enumerate}
\vspace*{-1mm}
\noindent However, even though opposites often occur in different
thesaurus categories, they can sometimes also be found in the same category. For example:
the word {\it ascent} is listed in the Macquarie Thesaurus categories of 49 ({\sc climbing}) and 694 ({\sc slope}), 
whereas the word {\it descent} is listed in the categories 40 ({\sc aristocracy}), 50 ({\sc dropping}), 538 ({\sc parentage}), and 694 ({\sc slope}).
Observe that {\it ascent} and {\it descent} are both
listed in the same category 694 ({\sc slope}),
which makes sense here because both words are pertinent to the concept of slope. 
On the other hand, two separate clues independently inform our system that the words are opposites of each other:
(1) Category 49 has the word {\it upwardness} in the same paragraph as {\it ascent},
and category 50 has the word {\it downwardness} in the same paragraph as {\it descent}.
The 13th affix pattern from Table \ref{tab:rules} ({\it up}X and {\it down}X) indicates that
the two thesaurus paragraphs have contrasting meaning. Thus, {\it ascent} and {\it descent}
occur in prime contrasting thesaurus paragraphs.
(2) One of the {\it ascent} categories (49) is adjacent to one of the {\it descent} categories (50),
and further this adjacent category pair has been manually marked as contrasting.

 Thus the words in a pair may be deemed both synonyms and opposites simultaneously by our methods of determining synonyms and opposites, respectively.
However, some of the features we use to determine opposites were found to be more precise (for example, words listed in
adjacent categories) than others (for example, categories identified as contrasting based
on affix and WordNet seeds).
Thus we apply the rules stated below 
as a decision list: if one rule fires, then the subsequent rules are ignored.
\vspace*{-2mm}
\begin{enumerate}
\item {\bf Rule 1 (high confidence for opposites):} If the words in a pair occur in adjacent thesaurus categories, then they are marked as opposites. 
\item {\bf Rule 2 (high confidence for synonyms):} If both the words in a pair occur in the same thesaurus category, then they are marked as synonyms. \\
\item {\bf Rule 3 (medium confidence for opposites):} If the words in a pair occur in prime contrasting thesaurus paragraphs, as determined by an affix-based or WordNet seed set, then they are marked as opposites. \\
\end{enumerate}
\vspace*{-1mm}

If a word pair is not tagged as synonym or opposite:
(a) the system can refrain from attempting an answer (this will attain high precision), 
or (b) the system can randomly guess the lexical relation (this will obtain 50\% accuracy for the pairs),
or (c) it could mark all remaining word pairs with the predominant lexical relation in the data (this will obtain an accuracy proportional to the skew in
distribution of opposites and synonyms).
For example, if after step 3, the system finds that 70\% of the marked word pairs were tagged opposites, and 30\% as
synonyms, then it could mark every hitherto untagged word pair (word pair for which it has insufficient information) as opposites.
We implemented all three variants.
Note that option (b) is indeed expected to perform poorly compared to option (c), but we include it
as part of our evaluation to measure usefulness of option (c).

\subsubsection{Results and discussion}
 Table~\ref{tab:Lin results} shows the precision (P), recall (R), and balanced F-score (F)  of various systems and baselines in
 identifying synonyms and opposites from the dataset described in Lin et al.\@ \shortcite{LinZQZ03}.
We will refer to this dataset as {\it LZQZ} (the first letters of the authors' last names).


\begin{table*}
    \caption{Results obtained on the synonym-or-opposite questions in LZQZ. 
	The best performing systems are marked in bold. The difference in precision and recall of method by Lin et al.\@ (2003) and our method in configurations b and c is not statistically significant.
	}
     \begin{center}
	{\small
    \begin{tabular}{lccc}
    \hline
						&P &R &F \\
	\hline 
	 	{\it Baselines:} 					&	&	&\\
		$\;\;\;$ a. random baseline 							&0.50 	&0.50 	&0.50 \\
		$\;\;\;$ b. supervised most-frequent baseline$^\dag$ 			&0.50 	&0.50 	&0.50 \\
	 	{\it Related work:} 										&	&	&\\
	 	$\;\;\;$ a. {\bf Lin et al.\@ (2003)}							&{\bf 0.90} 			&{\bf 0.90} 			&{\bf 0.90} \\
	 	$\;\;\;$ b. Turney (2011)							&0.82 			&0.82 			&0.82 \\
	 	{\bf \emph{Our method:}} if no information,  											&			&			&\\
	 	 $\;\;\;$ a. refrain from guessing		&0.98	&0.78		&0.87\\
	 	 $\;\;\;$ b. {\bf make random guess} 			&{\bf 0.88}		&{\bf 0.88}	&{\bf 0.88}\\
		 $\;\;\;$ c. {\bf mark the predominant class$^\ddag$} 		&{\bf 0.87}		&{\bf 0.87}		&{\bf 0.87}\\
	\hline
    \end{tabular}
	}
	\label{tab:Lin results}
     \end{center}

 	\normalsize
 	{\small $^\dag$This dataset has equal number of opposites and synonyms. Results reported are when choosing opposites as the predominant class.\\
 	$^\ddag$The system concluded that opposites were slightly more frequent than synonyms.} 
	\vspace*{-4mm}
\end{table*}

 If a system guesses at random (random baseline) it will obtain an accuracy of 50\%.
Choosing opposites (or synonyms) as the predominant class also obtains an accuracy of 50\% because the dataset 
has equal number of opposites and synonyms.
Published results on LZQZ \cite{LinZQZ03}  
are shown here again for convenience. 
The results obtained with our system
and the three variations on handling word pairs for which
it does not have enough information are shown in the last three rows.
The precision of our method in configuration a is significantly higher than that of Lin et al.\@ \shortcite{LinZQZ03}, with 95\% confidence according to the Fisher Exact Test \cite{Agresti90}.
Since precision and recall are the same for configuration b and c, as well as for the methods described in Lin et al.\@ \shortcite{LinZQZ03} and Turney \shortcite{Turney11},
we can also refer to these results simply as accuracy. We found that the differences in accuracies between the method of Lin et al.\@ \shortcite{LinZQZ03} 
and our method in configurations b and c are {\it not} statistically significant.
However, the method by Lin et al.\@ \shortcite{LinZQZ03} and our method in configuration b have significantly
higher accuracy than the method described in Turney \shortcite{Turney11}.
The lexical contrast features used in configurations a, b, and c, correspond to row i in Table \ref{tab:results}. 
The next subsection presents an analysis of the usefulness of the different features listed in Table \ref{tab:results}.

Observe that when our method refrains from guessing in case of insufficient information, it obtains excellent
precision (0.98), 
while still providing very good coverage
(0.78). 
As expected, the results obtained with (b) and (c) do not differ much from each other because the dataset has an equal number of synonyms and opposites.
(Note that the system was not privy to this information.)
However, after step 3 of the algorithm, the system had marked 65 pairs as opposites and 63 pairs as synonyms, and so it concluded that opposites are slightly more dominant in this dataset
and therefore the guess-predominant-class variant marked all previously unmarked pairs as opposites.

It should be noted that the LZQZ dataset was chosen
from a list of high-frequency terms. This was necessary to increase
the probability of finding sentences in a corpus where the target pair occurred in one of the chosen patterns proposed by Lin et al.\@ \shortcite{LinZQZ03}.
As shown in Table \ref{tab:results} earlier, the Lin et al.\@ (2003) patterns
have a very low coverage otherwise. Further the test data compiled by Lin et al.\@ 
only had opposites whereas the contrast questions had many contrasting word pairs that were not opposites. 

\begin{table*}
    \caption{Results obtained on the synonym-or-opposite questions in TURN. The best performing systems are marked in bold.}
     \begin{center}
	{\small
    \begin{tabular}{lccc}
    \hline
						&P &R &F \\
	\hline 
	 	{\it Baselines} 					&	&	&\\
		$\;\;\;$ a. random baseline 								&0.50 	&0.50 	&0.50\\
		$\;\;\;$ b. supervised most-frequent baseline$^\dag$ 		&0.65 	&0.65 	&0.65\\
	 	{\it Related work} 					&	&	&\\
	 	$\;\;\;$ a. Turney (2008)		&0.75	&0.75	&0.75\\
	 	$\;\;\;$ b. Lin et al.\@ (2003)		&0.35	&0.35	&0.35\\
	 	{\bf \emph{Our method:}} if no information, &	&	&\\
	 	 $\;\;\;$ a. refrain from guessing			&0.97	&0.69	&0.81\\
	 	 $\;\;\;$ b. {\bf make random guess} 				&{\bf 0.84}	&{\bf 0.84}	&{\bf 0.84}\\
		  $\;\;\;$ c. {\bf mark the predominant class$^\ddag$} 			&{\bf 0.90}	&{\bf 0.90}	&{\bf 0.90}\\

	\hline
    \end{tabular}
	}
	\label{tab:Turney results}
     \end{center}
 	\normalsize

	{\small $^\dag$About 65.4\% of the pairs in this dataset are opposites.
	So this row reports baseline results when choosing opposites as the predominant class.\\
	$^\ddag$The system concluded that opposites were much more frequent than synonyms.} 
	\vspace*{-4mm}
\end{table*}

 Table~\ref{tab:Turney results} shows results on the dataset described in Turney \shortcite{Turney08}.
We will refer to this dataset as {\it TURN}.
 The supervised baseline of always guessing the most frequent class (in this case, opposites), will obtain an
 accuracy of 65.4\% ($P$ = $R$ = $F$ = 0.654).

Turney \shortcite{Turney08} obtains an accuracy of 75\% using a supervised method and ten-fold
cross-validation.
A re-implementation of the method proposed by Lin et al.\@ \shortcite{LinZQZ03} as described earlier in Section 7.1.3
did not recognize any of the word pairs in TURN as opposites; that is, none of the word pairs in TURN
occurred in the Google n-gram corpus in patterns used by Lin et al.\@ \shortcite{LinZQZ03}. Thus it 
marked all words in TURN as synonyms.
The results obtained with our method are shown in the last three rows.
The precision and recall of our method in configurations b and c are significantly higher than those obtained by the methods by Turney \shortcite{Turney08} and Lin et al.\@ \shortcite{LinZQZ03}, with 95\% confidence according to the Fisher Exact Test \cite{Agresti90}.

Observe that once again our method, especially the variant that refrains from guessing in case of insufficient information, obtains excellent
precision (0.97), while still providing good coverage
(0.69). 
Also observe that results obtained by guessing the predominant class (method (c)) are markedly better 
than those obtained by randomly guessing in 
case of insufficient information (method (b)).
This is because, as mentioned earlier, the distribution of opposites and synonyms is somewhat skewed in this dataset (65.4\% of the pairs are opposites).
Of course, again the system was not privy to this information, but 
method (a) marked 58 pairs as opposites and 39 pairs as synonyms.
Therefore, the system concluded that opposites are more dominant 
and method (c) marked all previously unmarked pairs as opposites, obtaining an accuracy of 90\%.

Recall that in Section \ref{sec:ASmethod} we described how opposite pairs may occasionally be listed
in the same thesaurus category because the category may be pertinent to both words.
For 12 of the word pairs in the Lin et al.\@ data and 3 of the word pairs in the Turney data,
both words occurred together in the same thesaurus category, and yet the system
marked them as opposites because they occurred in adjacent thesaurus categories (Class I).
For 11 of the 12 pairs from LZQZ and for all 3 of the TURN pairs, this resulted in the correct answer.
These pairs are shown in Table~\ref{tab:syn2ants}.
 By contrast, only one of the term pairs in this table occurred in one of Lin's patterns of oppositeness, and was thus the only one correctly identified by their method as a pair of opposites.

It should also be noted that a word may have multiple meanings such that it may be synonymous to 
a word in one sense and opposite to it in another sense. Such pairs are also expected
to be marked as opposites by our system. Two such pairs in the
Turney (2008) data are: {\it fantastic}--{\it awful} and {\it terrific}--{\it terrible}.
The word {\it awful} can mean {\sc inspiring awe} (and so close to the meaning of {\it fantastic} in
some contexts), and also {\sc extremely disagreeable} (and so opposite to {\it fantastic}). 
The word {\it terrific} can mean {\sc frightful} (and so close to the meaning of {\it terrible}),
and also {\sc unusually fine} (and so opposite to {\it terrible}).
Such pairs are probably not the best synonym-or-opposite questions. 
However, faced with these questions,
humans probably home in on the dominant senses of the target words to determine an answer. 
For example,
in modern-day English {\it terrific} is used more frequently in the sense of {\sc unusually fine}
than the sense of {\sc frightful}, and so most people will say that {\it terrific} and
{\it terrible} are opposites (in fact that is the solution provided with this data).

\begin{table*}
    \caption{Pairs from LZQZ and TURN that have at least one category in common but are still marked as opposites by our method.}
     \centering
	{\small
    \begin{tabular}{llc cc llc}
	\cline{0-2} \cline{6-8}
						\multicolumn{3}{c}{\bf LZQZ} & &  &\multicolumn{3}{c}{\bf TURN}\\
						{\bf word 1} &{\bf word 2} &{\bf official solution} & & 	&{\bf word 1} &{\bf word 2} &{\bf official solution}\\
	\cline{0-2} \cline{6-8}
					amateur 		&professional	&opposite	& &		&fantastic 		&awful	&opposite\\
					ascent 		&descent	&opposite			& &		&dry 		&wet	&opposite\\
					back 		&front	&opposite				& &		&terrific 		&terrible	&opposite\\ \cline{6-8}

					bottom 		&top	&opposite				& &		& & &\\
					broadside 		&salvo	&synonym		& &		& & &\\
					entrance 		&exit	& opposite		& &		& & &\\
					heaven 		&hell	&opposite			& &		& & &\\
					inside 		&outside	&opposite		& &		& & &\\
					junior 		&senior	&opposite			& &		& & &\\
					lie 		&truth	&opposite			& &		& & &\\
					majority 		&minority	&opposite		& &		& & &\\
					nadir 		&zenith	&opposite			& &		& & &\\
					strength 		&weakness	&opposite	& &		& & &\\
	\cline{0-2} 
	\end{tabular}
	}
	\label{tab:syn2ants}
 	\normalsize
	\vspace*{-4mm}
\end{table*}

\subsubsection{Analysis} 

We carried out additional experiments to determine how useful individual components of our method were in
solving the synonym-or-opposite questions. The results on 
LZQZ are shown in Table \ref{tab:FA Lin} and the results on TURN are shown in
Table \ref{tab:FA Turney}. These results are for the case when the system refrains from guessing in case of
insufficient information. The rows in the tables correspond to the rows in Table \ref{tab:results}
shown earlier that gave results on the contrast questions.

\begin{table*}
    \caption{Results for individual components as well as certain combinations of components on the synonym-or-opposite questions in LZQZ. 
	The best performing configuration is shown in bold.}
     \begin{center}
	  {\small
    \begin{tabular}{l ccc }
    \hline
						&P &R &F \\
	\hline 
	{\it Baselines:}											&   &   & \\
	$\;\;\;$ a. random baseline					 			&0.50   &0.50   &0.50 \\
    $\;\;\;$ b. supervised most-frequent baseline$^\dag$       &0.50   &0.50   &0.50 \\
	{\it Our methods:}											&   &   & \\
	$\;\;\;$ a. affix-generated seeds only							&0.86	&0.54    &0.66  \\
	$\;\;\;$ b. WordNet seeds only 									&0.88	&0.65    &0.75  \\
	$\;\;\;$ c. both seed sets (a + b)	 							&0.88	&0.65    &0.75  \\
	$\;\;\;$ d. adjacency heuristic only								&0.95	&0.74    &0.83 \\
	$\;\;\;$ e. manual annotation of adjacent categories 						&0.98	&0.74    &0.84 \\
	$\;\;\;$ f. affix seed set and adjacency heuristic (a + d)						&0.95	&0.75    &0.84 \\
	$\;\;\;$ g. both seed sets and adjacency heuristic (a + b + d) 					 	&0.95	&0.78    &0.86 \\
	$\;\;\;$ h. affix seed set and annotation of adjacent categories						&0.98	&0.77    &0.86 \\
	 $\;\;\;\;\;\;$ (a + e)					&   &   & \\
	$\;\;\;$ i. {\bf both seed sets and annotation of adjacent categories}					&{\bf 0.98} &{\bf 0.78}      &{\bf 0.87} \\
	 $\;\;\;\;\;\;$ {\bf (a + b + e)}				&   &   & \\
	\hline
    \end{tabular}
	}
	\label{tab:FA Lin}
     \end{center}
 	\normalsize

	{\small $^\dag$ This dataset has equal number of opposites and synonyms, so either class can be chosen to be predominant.
	    Baseline results shown here are for choosing opposites as the predominant class.}
 \vspace{-5mm}
\end{table*}

\begin{table*}
    \caption{Results for individual components as well as certain combinations of components on the synonym-or-opposite questions in TURN. 
	The best performing configuration is shown in bold.}
      \begin{center}
	  {\small

    \begin{tabular}{l ccc }
    \hline
						&P &R &F \\
	\hline 
	{\it Baselines:}											&   &   &   \\
	$\;\;\;$ a. random baseline					 			&0.50   &0.50    &0.50 \\
    $\;\;\;$ b. supervised most-frequent baseline$^\dag$       &0.65   &0.65    &0.65  \\
	{\it Our methods:}											&   &   & \\
	$\;\;\;$ a. affix-generated seeds only							&0.92	&0.54    &0.68  \\
	$\;\;\;$ b. WordNet seeds only 									&0.93	&0.61    &0.74  \\
	$\;\;\;$ c. both seed sets (a + b)	 							&0.93	&0.61    &0.74  \\
	$\;\;\;$ d. adjacency heuristic only								&0.94	&0.60    &0.74  \\
	$\;\;\;$ e. manual annotation of adjacent categories  						&0.96	&0.60    &0.74 \\
	$\;\;\;$ f. affix seed set and adjacency heuristic (a + d) 						&0.95	&0.67    &0.78 \\
	$\;\;\;$ g. both seed sets and adjacency heuristic (a + b + d) 						&0.95	&0.68    &0.79 \\
	$\;\;\;$ h. affix seeds and annotation of adjacent categories 						&0.97	&0.68    &0.80 \\
	 $\;\;\;\;\;\;$ (a + e)					&   &   & \\
	$\;\;\;$ i. {\bf both seed sets and annotation of adjacent  categories}					&{\bf 0.97} &{\bf 0.69}   &{\bf 0.81} \\
	 $\;\;\;\;\;\;$ {\bf (a + b + e)}				&   &   & \\
	\hline
    \end{tabular}
	}
	\label{tab:FA Turney}
      \end{center}
 	\normalsize

	{\small $^\dag$ About 65.4\% of the pairs in this dataset are opposites.
	    So this row reports baseline results when choosing opposites as the predominant class.}
\vspace{-5mm}
\end{table*}


Observe that the affix-generated seeds give a marked improvement over the baselines, and that knowing which categories
are contrasting (either from the adjacency heuristic or manual annotation of adjacent categories)
proves to be the most useful feature. Also note that even though manual annotation and WordNet seeds eventually
lead to the best results (F = 0.87 for LZQZ and F = 0.81 for TURN), using only the adjacency 
heuristic and the affix-generated seeds gives competitive results (F = 0.84 for the Lin set and F = 0.78 for the Turney set).
We are interested in developing methods to make the approach
cross-lingual, so that we can use a thesaurus from one language (say English) to compute lexical contrast
in a resource-poor target language.

The precision of our method is very good (> 0.95). Thus future work will be aimed at improving recall.
This can be achieved by developing methods to generate more seed opposites. This is also an avenue
through which some of the pattern-based approaches (such as the methods described by Lin et al.\@ (2003) and Turney (2008)) can be incorporated
into our method. For instance, we could use n-gram patterns such as 
``either X or Y'' and ``from X to Y'' to identify
pairs of opposites that can be used as additional seeds in our method. 

Recall can also be improved by using affix patterns in other languages to identify contrasting thesaurus
paragraphs in the target language. Thus, constructing a cross-lingual framework in which words from one language
will be connected to thesaurus categories in another language will be useful not only in
computing lexical contrast in a resource-poor language, but also in using affix information from different
languages to improve results in the target, possibly even resource-rich, language.

\section{Conclusions and Future Work}
\label{sec:conclusion}

 Detecting semantically contrasting word pairs
has many applications in natural language processing.
In this paper, we proposed a method for computing lexical contrast that is based on the 
hypothesis that 
if a pair of words, $A$ and $B$, are contrasting, then there is a pair of opposites, $C$ and $D$, such that $A$ and $C$ are strongly related and $B$ and $D$ are strongly related---the contrast hypothesis. 
We used pointwise mutual information to determine the degree of contrast between two contrasting words.
The method outperformed others on the task of solving a large set of ``choose the most contrasting word'' questions
wherein the system not only identified whether two words are contrasting
but also distinguished between pairs of contrasting words with differing degrees of contrast.
We further determined performance of the method on five different kinds of opposites and across four parts of speech.
We used our approach to solve synonym-or-opposite questions described in Turney (2008) and Lin et al.\@ (2003).

 Since opposites were central to our methodology, we 
designed a questionnaire to better understand different kinds of opposites,
which we crowdsourced with Amazon Mechanical Turk.
We devoted extra effort to making sure the questions are phrased in a simple, yet clear manner.
Additionally,  a quality control method was developed, using a
word-choice question, to automatically identify and discard  dubious and outlier annotations. 
From this data, we created a dataset of different kinds of opposites that we have made available. 
We determined the amount of agreement among humans in identifying lexical contrast,
and also in identifying different kinds of opposites. 
We also 
showed that a large number of opposing word pairs have properties pertaining to more than one kind.
Table~\ref{tab:data} summarizes the data created as part of this research on lexical contrast,
all of which is available for download. 

 New questions that target other types of lexical contrast not addressed in this paper may be added in the future. 
It may be desirable to break a complex question into two or more simpler questions. For example, if a word pair is considered to be a certain kind of opposite when it has both properties $M$ and $N$, 
then it is best to have two separate questions asking whether the word pair has properties $M$ and $N$, respectively.
The crowdsourcing study can be replicated for other languages by asking the same questions in the target language for words in the target language. 
Note, however, that as of February 2012, most of the Mechanical Turk participants are native speakers of English, certain Indian languages, and some European languages.

\begin{table}[]
\caption{A summary of the data created as part of this research on lexical contrast. Available for download at: {\protect \url {http://www.purl.org/net/saif.mohammad/research.}}}
\centering
 {\small
\begin{tabular}{l l}
\hline
      Name    &\# of items\\
\hline
Affix patterns that tend to generate opposites &15 rules\\[1mm]
Contrast questions: &\\
$\;\;\;\;\;\;\;$ GRE preparatory questions: &\\
$\;\;\;\;\;\;\;\;\;\;\;\;\;$ Development set &162 questions\\ 
$\;\;\;\;\;\;\;\;\;\;\;\;\;$ Test set &790 questions\\
$\;\;\;\;\;\;\;$ Newly created questions: &1269 questions\\[1mm]
Data from work on types of opposites: &\\
$\;\;\;\;\;\;\;$ Crowdsourced questionnaires  &4 sets (one for every pos)\\
$\;\;\;\;\;\;\;$ Responses to questionnaires & 12,448 assignments (in four files)\\[1mm]
Lexicon of opposites generated by the &\\
Mohammad et al. method: &\\
$\;\;\;\;\;\;\;$ Class I opposites  &3.5 million word pairs\\
$\;\;\;\;\;\;\;$ Class II opposites &2.5 million word pairs\\[1mm]
Manually identified contrasting categories &\\
in the Macquarie Thesaurus &209 category pairs\\[1mm]
Word-pairs used in Section 5 experiments:&\\
$\;\;\;\;\;\;\;$ WordNet opposites set  &1358 word pairs\\
$\;\;\;\;\;\;\;$ WordNet random word pairs set &1358 word pairs\\
$\;\;\;\;\;\;\;$ WordNet synonyms set &1358 word pairs\\
\hline
\end{tabular}                   
 }
\label{tab:data}
\vspace{-5mm}
\end{table}

Our future goals include porting this approach to a cross-lingual framework
 to determine lexical contrast in  a resource-poor language by using a bilingual lexicon
 to connect the words in that language with words in another resource-rich language.
We can then use the structure of the thesaurus from the resource-rich language as described in this paper to detect contrasting categories of terms. 
This is similar to the approach described by Mohammad et al.\@ \shortcite{MohammadGHZ07}, who
compute semantic distance in a resource-poor language by using a bilingual lexicon and
a sense disambiguation algorithm to connect
text in the resource-poor language with a thesaurus in a different language.
This enables automatic discovery of lexical contrast in a language
even if it does not have a Roget-like thesaurus. However, the cross-lingual method
still requires a bilingual lexicon to map words between the target language
and the language with the thesaurus.

Our method used only one Roget-like published thesaurus, but even more gains
may be obtained by combining many dictionaries and thesauri using methods
proposed by Ploux and Victorri \shortcite{PlouxV98} and others.\\

We modified our algorithm to create lexicons of words
associated with positive and negative sentiment \cite{MohammadDD09}.
 We also used the lexical contrast algorithm in some preliminary experiments to identify contrast between sentences and use that information to improve
 cohesion in automatic summarization \cite{UMDTAC2008}.
Since its release, the lexicon of contrasting word pairs was used to improve textual paraphrasing and in turn help improve machine translation \cite{MartonWMT11}.
We are interested in using contrasting word pairs as seeds to identify phrases
that convey contrasting meaning. These will be especially helpful in machine translation 
where current systems have difficulty separating translation hypotheses that convey
the same meaning as the source sentences, and those that do not.

 Given a particular word, our method computes a single score as the degree of contrast with another another.
However, a word may be more or less contrasting with another word, when used in different contexts \cite{Murphy03}.
Just as in the lexical substitution task \cite{McCarthy09}, where a system has to find the word that can
best replace a target word in context to preserve meaning, one can imagine a lexical substitution task to generate
contradictions where the objective is to replace a given target word with one that is contrasting so as to
generate a contradiction. 
Our future work includes developing a context-sensitive measure of lexical contrast
that can be used for exactly such a task.

There is considerable evidence that children are aware of lexical contrast at a very early age \cite{Murphy08}.
They rely on it to better understand various concepts
and in order to communicate effectively. 
Thus we believe that computer algorithms that deal with language can also obtain significant gains
through the ability to detect contrast and the ability to distinguish between differing degrees of contrast. 


\section*{Acknowledgments}

We thank Tara Small, Smaranda Muresan, and Siddharth Patwardhan
for their valuable feedback.  This work was supported
in part by the National Research Council Canada,
in part by the National Science Foundation under Grant No. IIS-0705832,
in part by the Human Language Technology Center of Excellence, and in
part by the Natural Sciences and Engineering Research Council of
Canada.  Any opinions, findings, and conclusions or recommendations
expressed in this material are those of the authors and do not
necessarily reflect the views of the sponsors.

\bibliographystyle{fullname}
\bibliography{references}

\end{document}